%% file: main.tex
\documentclass[10pt,twocolumn,letterpaper]{article}

\usepackage{iccv}              % To produce the CAMERA-READY version

\definecolor{iccvblue}{rgb}{0.21,0.49,0.74}
\usepackage[pagebackref,breaklinks,colorlinks,allcolors=iccvblue]{hyperref}
\usepackage{multirow}

\def\paperID{4451} % *** Enter the Paper ID here
\def\confName{ICCV}
\def\confYear{2025}

\title{Instant GaussianImage: A Generalizable and Self-Adaptive Image Representation via 2D Gaussian Splatting}

\author{\textbf{Zhaojie Zeng$^{1}$, Yuesong Wang$^{1}\thanks{Corresponding author.}$, Chao Yang$^{2}$, Tao Guan$^{1}$, Lili Ju$^{3}$}\\
$^{1}$ School of Computer Science \& Technology, Huazhong University of Science and Technology\\
$^{2}$ NERCGIS, China University of Geoscience (Wuhan)\\
$^{3}$ Department of Mathematics, University of South Carolina\\
{\tt\small zhaojiezeng@hust.edu.cn, yuesongwang@hust.edu.cn}
}

\begin{document}
\maketitle
\input{sec/0_abstract}
\input{sec/1_intro}
\input{sec/2_related}

\input{sec/3_method}
\input{sec/4_experiment}
\input{sec/5_conclusion}
\input{sec/6_acknowledge}
{
    \small
    \bibliographystyle{ieeenat_fullname}
    \bibliography{main}
}

\end{document}

\maketitle

In this supplementary material, we provide additional explanations for conclusions presented in the main paper, illustrate more technical details, and supplement experimental results.

\section{Explanations}
% 1. 提供 CDF 采样会导致点云分布不合理的证明
\subsection{Analysis of CDF Sampling in \texorpdfstring{Sec.~3.3}{Sec. 3.3}}

In Sec.~3.3 of the main paper (\textit{Dithering with} $\mathbb{P}_\text{pred}$), we state:  
 \begin{quote}
    ``An alternative is CDF sampling with a fixed number of Gaussians, as in \cite{ImageGS}. While it works when high- and low-entropy regions are balanced, it degrades rendering quality when high-entropy regions dominate, reducing point density where detail is needed. Conversely, in low-entropy images, excessive points are allocated to smooth regions, leading to redundancy."
\end{quote}

This implies that when an image is dominated by low-entropy or high-entropy regions, the CDF-based sampling strategy also suffers from overfitting or underfitting issues\cite{3DGS}, similar to the threshold-based sampling.

\begin{figure*}[ht!]
\begin{center}
\includegraphics[width=\linewidth]{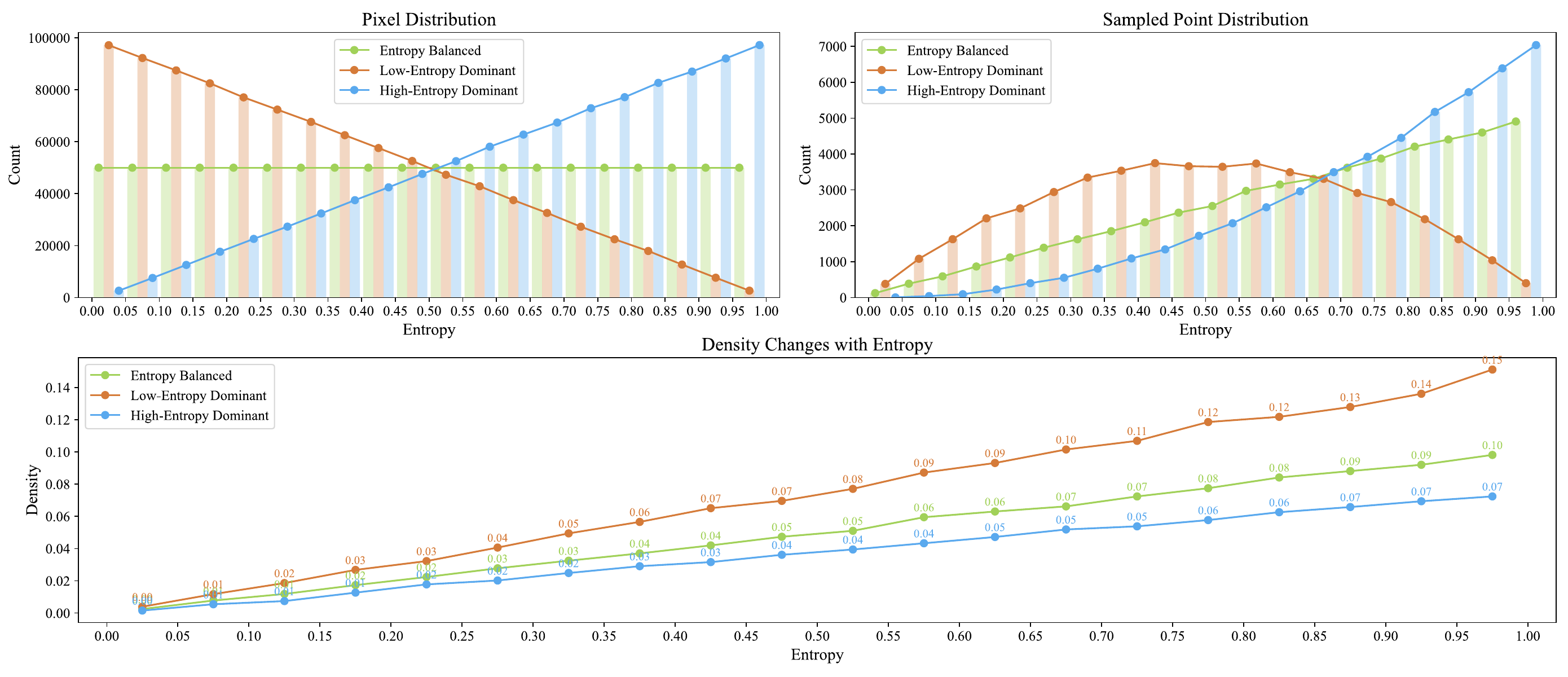}
\vspace{-1.0cm}
\end{center}
   \caption{\noindent \textbf{CDF Sampling Analysis.} A simulated experiment analyzing the effects of CDF sampling under different entropy distributions. \textbf{Top left}: Pixel distribution for balanced, low-entropy dominant, and high-entropy dominant cases. \textbf{Top right}: Sampled point distribution after applying CDF sampling. \textbf{Bottom}: Normalized sampling density, showing redundancy in low-entropy dominant cases (overfitting) and insufficient representation in high-entropy dominant cases (underfitting).
}
\label{fig:SampledPointDistribution}
\end{figure*}

To further analyze the behavior of CDF sampling, we conduct a simple simulation experiment. We generate three types of synthetic image data with different entropy distributions:  
\begin{itemize}
    \item \textbf{Balanced entropy} (green) – where pixel counts are evenly distributed across entropy intervals.  
    \item \textbf{Low-entropy dominant} (orange) – where low-entropy regions contain more pixels.  
    \item \textbf{High-entropy dominant} (blue) – where high-entropy regions contain more pixels.  
\end{itemize}

As shown in \cref{fig:SampledPointDistribution} the \textbf{top left} plot, these distributions define the initial pixel counts per entropy interval. We then apply CDF sampling to each distribution and obtain the number of sampled points in each entropy interval, illustrated in the \textbf{top right} plot. It is evident that in both the balanced and high-entropy dominant cases, more points are allocated to high-entropy regions, while in the low-entropy dominant case, mid-entropy regions receive the highest number of sampled points.

To quantify the sampling efficiency, we normalize the number of sampled points by the corresponding pixel counts in each entropy interval, resulting in the \textbf{bottom plot}. Using the balanced case as a reference, we observe that:  
\begin{itemize}
    \item When low-entropy regions dominate, the sampling density is significantly higher than the balanced case, leading to redundancy (\textit{overfitting}). 
    \item When high-entropy regions dominate, the sampling density is notably lower than the balanced case, leading to insufficient representation (\textit{underfitting}).
\end{itemize}
These results confirm that CDF sampling alone struggles to maintain an optimal balance between high- and low-entropy regions.

\subsection{Rationale Behind Using Delaunay Triangulation in \texorpdfstring{Sec.~3.4}{Sec. 3.4}}  

In Sec.~3.4 \textit{Ellipse Fitting}, we argue against directly predicting the scaling of each Gaussian using a network. Instead, we first perform Delaunay triangulation and then predict the offsets. The primary reasons for this choice are as follows:  

Most deep learning methods handling similar tasks~\cite{Scaffold-GS} adopt a normalized scale approach, such as learning relative sizes or operating in a specific feature space to enhance adaptability. While some studies directly predict pixel-level absolute values, such methods typically rely on fixed image sizes. If the input dimensions vary, these predictions may become invalid. Given that our task involves images of varying sizes, a normalized scale approach is necessary.  

Several normalization strategies exist. One simple approach is to define an absolute scale based on the nearest neighbor distance and then learn a relative scale accordingly. While effective for uniformly distributed points, this method struggles with non-uniform distributions—closely spaced points may result in an absolute scale that fails to provide full coverage, increasing training complexity.  

Thus, we prioritize an initialization strategy that ensures full image coverage while minimizing overlap between primitives to reduce learning complexity. To achieve this, we adopt \textbf{Delaunay Triangulation}.  

Specifically, after obtaining the points via Floyd-Steinberg Dithering, we compute the average nearest-neighbor distance \( \delta \). To ensure full coverage, we insert additional boundary points at intervals of approximately \( 3\delta \) along the image edges. After triangulation, we fit ellipses to the generated triangles using OpenCV’s \texttt{fitEllipse} function~\cite{fitEllipse}, which requires at least six discrete points. To meet this requirement, we insert the midpoints of each triangle’s three edges, yielding six points per triangle.  

The fitted major and minor axes are scaled by \( 0.5 \) to serve as the absolute scale of each Gaussian. The fitted ellipse centers and rotation angles are normalized to the range \([-1,1]\).

\section{Details}

\begin{figure}[ht!]
\begin{center}
\includegraphics[width=\linewidth]{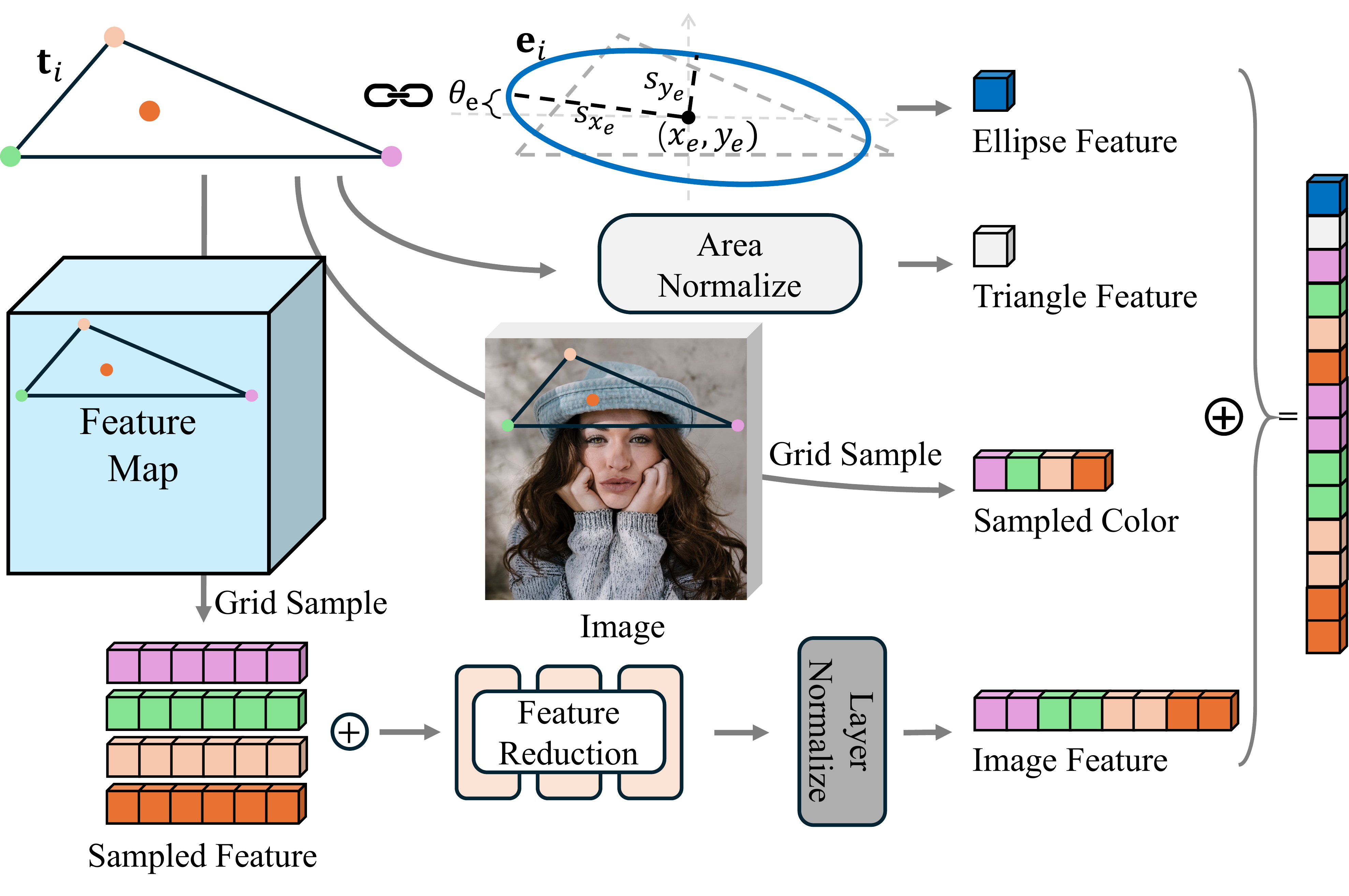}
\end{center}
   \caption{
   \textbf{Feature organization for MLP input.} We construct the MLP input by combining multiple feature components. Geometric features are extracted from the triangle \( \mathbf{t}_i \) and its fitted ellipse \( \mathbf{e}_i \), while grid sampling provides local deep features and sampled colors from the input image. The extracted features undergo feature reduction and layer normalization before concatenation, forming the final input vector for the MLP.
    }
\label{fig:feature_organization}
\end{figure}

\subsection{Feature Extraction Details}
Due to space limitations in the main paper, we provide a detailed explanation of how the features in Fig.~\ref{fig:feature_organization} are obtained. These include the \textit{Triangle Feature}, \textit{Ellipse Feature}, \textit{Sampled Color}, and \textit{Image Feature}. After performing Delaunay triangulation, we obtain a set of triangle representations \( \{\mathbf{t}_i\}_{i=0}^{T} \), where each \( \mathbf{t}_i \) consists of the three vertex coordinates of the corresponding triangle:
\begin{equation}
    \mathbf{t}_i = [\mathbf{v}_a, \mathbf{v}_b, \mathbf{v}_c]^T = \begin{bmatrix} x_a & y_a \\ x_b & y_b \\ x_c & y_c \end{bmatrix},
\end{equation}

\noindent \textbf{Triangle Feature.}  
Our goal is to normalize the triangle's area and obtain relative coordinates in a normalized space. Given a set of triangles represented by their three vertex coordinates $\mathbf{t}_i$, we first compute the area of each triangle as:
\begin{equation}
    A_i = \frac{1}{2} \left| x_a(y_b - y_c) + x_b(y_c - y_a) + x_c(y_a - y_b) \right|.
\end{equation}
To normalize the scale, we apply a transformation:
\begin{equation}
    s_i = \sqrt{\frac{1}{A_i}},
\end{equation}
where \( s_i \) is the scaling factor. We then compute the triangle center:
\begin{equation}
    \mathbf{c}_i = \frac{1}{3} ( \mathbf{v}_a + \mathbf{v}_b + \mathbf{v}_c ),
\end{equation}
where \( \mathbf{v}_a, \mathbf{v}_b, \mathbf{v}_c \) are the vertex coordinates. The final normalized triangle representation is given by:
\begin{equation}
    \mathbf{t}_i^{t} = s_i (\mathbf{t}_i - \mathbf{c}_i),
\end{equation}
which ensures that all triangles have a consistent scale while preserving their relative shape. The obtained normalized coordinates \( \mathbf{t}_i^{t} \) serve directly as our triangle feature (dimension is 6).

\noindent \textbf{Ellipse Feature.}  
We adopt the ellipse fitting method described in~\cite{fitEllipse}, which requires at least six points. For each triangle, we construct a vertex set:
\begin{equation}
    \mathbf{t}_i^{e} = \{  \mathbf{v}_a, \mathbf{v}_b, \mathbf{v}_c, \frac{\mathbf{v}_a + \mathbf{v}_b}{2}, \frac{\mathbf{v}_b + \mathbf{v}_c}{2}, \frac{\mathbf{v}_a + \mathbf{v}_c}{2} \}.
\end{equation}
Using these points, we fit an ellipse and obtain a set of ellipses \( \{\mathbf{e}_i\}_{i=0}^{T} \), where each ellipse is characterized by its center position, major and minor axes, and rotation angle:
\begin{equation}
    \mathbf{e}_i = \{(x_e, y_e), (s_{x_e}, s_{y_e}), \theta_{e} \}.
\end{equation}
The final ellipse feature (dimension is 4) is represented as:
\begin{equation}
    \mathbf{f}_i^{e} = \{ x_e, y_e, \frac{s_{x_e}}{s_{y_e} + 10^{-6}}, e_{\theta} \}.
\end{equation}

\noindent \textbf{Color Feature}  
To extract the color feature, we sample colors from the image at the three triangle vertices and the triangle center. The sampling positions are defined as:
\begin{equation}
    \mathbf{t}_i^{c} = \{ \mathbf{v}_a, \mathbf{v}_b, \mathbf{v}_c, \mathbf{c}_i \},
\end{equation}
where the triangle center is computed as:
\begin{equation}
    \mathbf{c}_i = \frac{1}{3} (\mathbf{v}_a + \mathbf{v}_b + \mathbf{v}_c).
\end{equation}
The color values at these positions are obtained using \textbf{grid sampling} on image $I$, which served directly as our color feature (dimension is 12).

\begin{equation}
    \mathbf{f}_i^{c} = I(\mathbf{t}_i^{c}).
\end{equation}

\noindent \textbf{Image Feature}  
Similar to the color feature extraction, we sample features from the 64-dimensional feature map produced by the ConvNeXt-based UNet at the three triangle vertices and the triangle center:
\begin{equation}
    \mathbf{t}_i^{f} =\mathbf{t}_i^{c}.
\end{equation}
The feature values at these positions are obtained using \textbf{grid sampling} on feature map $F$:
\begin{equation}
    \mathbf{f}_i^{f} = F(\mathbf{t}_i^{f}).
\end{equation}
The sampled features are then concatenated and processed through a feature reduction network:
\begin{equation}
    \mathbf{f}_i^{r} = \text{ReLU}(\mathbf{W}_3 \text{ReLU}(\mathbf{W}_2 \text{ReLU}(\mathbf{W}_1 \mathbf{f}_i))),
\end{equation}
where \( \mathbf{W}_1, \mathbf{W}_2, \mathbf{W}_3 \) are linear transformation matrices reducing the feature dimension sequentially from \( 4d \to 4d \to 2d \to d \). Finally, the output feature is normalized using Layer Normalization. The final image feature dimension is $d=64$.

\subsection{Network Design}
\noindent \textbf{ConvNeXt-based UNet.}  
For feature extraction, we employ a ConvNeXt-based UNet. The encoder uses the \texttt{base} model configuration with:
\begin{equation}
    \text{depths} = [3, 3, 27, 3], \quad \text{dims} = [128, 256, 512, 1024].
\end{equation}
The decoder follows the implementation of Pytorch-UNet.

\noindent \textbf{Position Field.} 
The Position Field network transforms the feature of each pixel into a probability value to generate the Position Probability Map (PPM). It takes a 64-dimensional feature vector as input and outputs a single probability value:
\begin{equation}
    \mathbf{p}_\text{out} = \sigma(\mathbf{W}_3 \text{ReLU}(\mathbf{W}_2 \text{ReLU}(\mathbf{W}_1 \mathbf{p}_\text{in}))),
\end{equation}
where \( \mathbf{p}_\text{in} \in \mathbb{R}^{64} \) is the input feature, \( \mathbf{W}_1, \mathbf{W}_2, \mathbf{W}_3 \) are linear transformation matrices, and \( \sigma \) denotes the Sigmoid activation function to ensure output values are within \([0,1]\).

\noindent \textbf{BC Field.}  
The BC Field network transforms the feature of each primitive (\(\text{dim} = 86\)) into barycentric coordinates (\(\text{dim} = 3\)):
\begin{equation}
    \mathbf{bc}_\text{out} = \text{Softmax}(\mathbf{W}_3 \text{ReLU}(\mathbf{W}_2 \text{ReLU}(\mathbf{W}_1 \mathbf{b}_\text{in})),
\end{equation}
where \( \mathbf{bc}_\text{in} \in \mathbb{R}^{86} \) is the input feature, \( \mathbf{W}_1, \mathbf{W}_2, \mathbf{W}_3 \) are linear transformation matrices, and the Softmax activation ensures that the output barycentric coordinates sum to 1.

\noindent \textbf{$\Sigma$ Field.}  
The $\Sigma$ Field predicts offsets for scaling and rotation, adjusting the major and minor axes as well as the orientation of the fitted ellipse. It takes a 94-dimensional feature vector as input and outputs a 3-dimensional transformation parameter:
\begin{equation}
    \mathbf{\Sigma}_\text{out} = \mathbf{W}_3 \text{ReLU}(\mathbf{W}_2 \text{ReLU}(\mathbf{W}_1 \mathbf{\Sigma}_\text{in})),
\end{equation}
where \( \mathbf{s}_\text{in} \in \mathbb{R}^{86} \) represents the input feature, and \( \mathbf{W}_1, \mathbf{W}_2, \mathbf{W}_3 \) are linear transformation matrices. The output consists of two scaling factors for the major and minor axes and one rotation offset.

\noindent \textbf{Opacity Field.} 
The Opacity Field predicts the opacity of each primitive based on its feature representation. It takes a 94-dimensional input feature and outputs a single scalar value:
\begin{equation}
    o_\text{out} = \sigma(\mathbf{W}_3 \text{ReLU}(\mathbf{W}_2 \text{ReLU}(\mathbf{W}_1 \mathbf{o}_\text{in}))),
\end{equation}
where \( \mathbf{o}_\text{in} \in \mathbb{R}^{86} \) is the input feature, \( \mathbf{W}_1, \mathbf{W}_2, \mathbf{W}_3 \) are linear transformation matrices, and \( \sigma \) (Sigmoid) ensures the output opacity value is within the range \([0,1]\).

\subsection{Obtain Pseudo PPM.}  
To train a network capable of predicting the Position Probability Map (PPM) that represents the Gaussian distribution, we first generate high-quality training data via Gaussian decomposition using GaussianImage.  We begin with \textit{quadtree-based image partitioning}, where blocks are recursively subdivided based on the mean squared error (MSE) of colors. If the MSE exceeds a predefined threshold, the block is split into four smaller ones, continuing until either the minimum block size of \( 4 \times 4 \) is reached or the MSE falls below \( 0.02 \). This adaptive partitioning can approximately estimate the required number of Gaussians for good image representation, thus preventing the generation of a low-quality pseudo PPM due to the inadequate number of Gaussians. Next, Gaussians are initialized at the center of each block. Finally, we train GaussianImage for 50,000 iterations.

After Gaussian decomposition, we compute the Gaussian density at each pixel position \(\mathbf{x}\) in the image. Specifically, for each \(\mathbf{x}\), we estimate the minimal radius of a circle  that encompasses \( K \) Gaussians, then the local Gaussian density $D_{\mathbf{x}}$ is defined as the unit density of Gaussians in this circle:
\begin{equation}
    D_{\mathbf{x}} =  \frac{K}{\pi \cdot {\max\limits_{i} \left( \|\mathbf{p}_{\mathbf{x}} - \mathbf{p}_i\| \right) }^2},
\end{equation}
where \(\mathbf{p}\) represents position coordinates, and \( i \) indexes the top-\( K \) nearest neighbors of \(\mathbf{x}\). In our experiments, we set \( K = 10 \).  
Next, we convert this density map into a PPM, which determines the probability of generating a Gaussian at each pixel location:
\begin{equation}
    \mathbb{P}_{\text{pseudo}}(\mathbf{x}) = f(D_{\mathbf{x}}) = \mathcal{N}_{3\sigma} \left( \frac{K}{{\max\limits_{i} \left( \|\mathbf{p}_{\mathbf{x}} - \mathbf{p}_i\| \right)^{1/2} }} \right),
\end{equation}
where $f$ is a mapping function for a more smooth distribution and \( \mathcal{N}_{3\sigma} \) denotes Three-Sigma Clipped Normalization, formally defined as:
\begin{equation}
\begin{aligned}
    \mu &= \text{mean}(\mathbf{x}) \\
    \sigma &= \text{std}(\mathbf{x}) \\
    x_{\max} &= \min(\mu + 3\sigma, \max(\mathbf{x})) \\
    x_{\min} &= \max(\mu - 3\sigma, \min(\mathbf{x})) \\
    x' &= \frac{x - x_{\min}}{x_{\max} - x_{\min}} \\
    x_{\text{normalized}} &= \text{clamp}(x', 0, 1)
\end{aligned}
\end{equation}

\subsection{Dithering on PPM.}
A key aspect of our method is the discretization of the predicted PPM using Floyd-Steinberg Dithering. Direct per-pixel processing would generate an excessive number of sampled points; therefore, we apply \textit{max pooling} to divide the PPM into \( k \times k \) patches, where each patch's probability value is determined by the maximum probability within the patch.  By adjusting \( k \), we control the number of generated Gaussians to achieve different levels of rendering detail.For example, \( k=3 \) for high-detail rendering while \( k=4 \) for balanced rendering. To accelerate dithering, we implement a GPU-accelerated version based on~\cite{DitheringGPU}. After obtaining the sampled points, we apply \textit{upsampling} to restore their positions to the original resolution.

\section{More Experiments}

\begin{figure*}[ht!]
\begin{center}
\includegraphics[width=\linewidth]{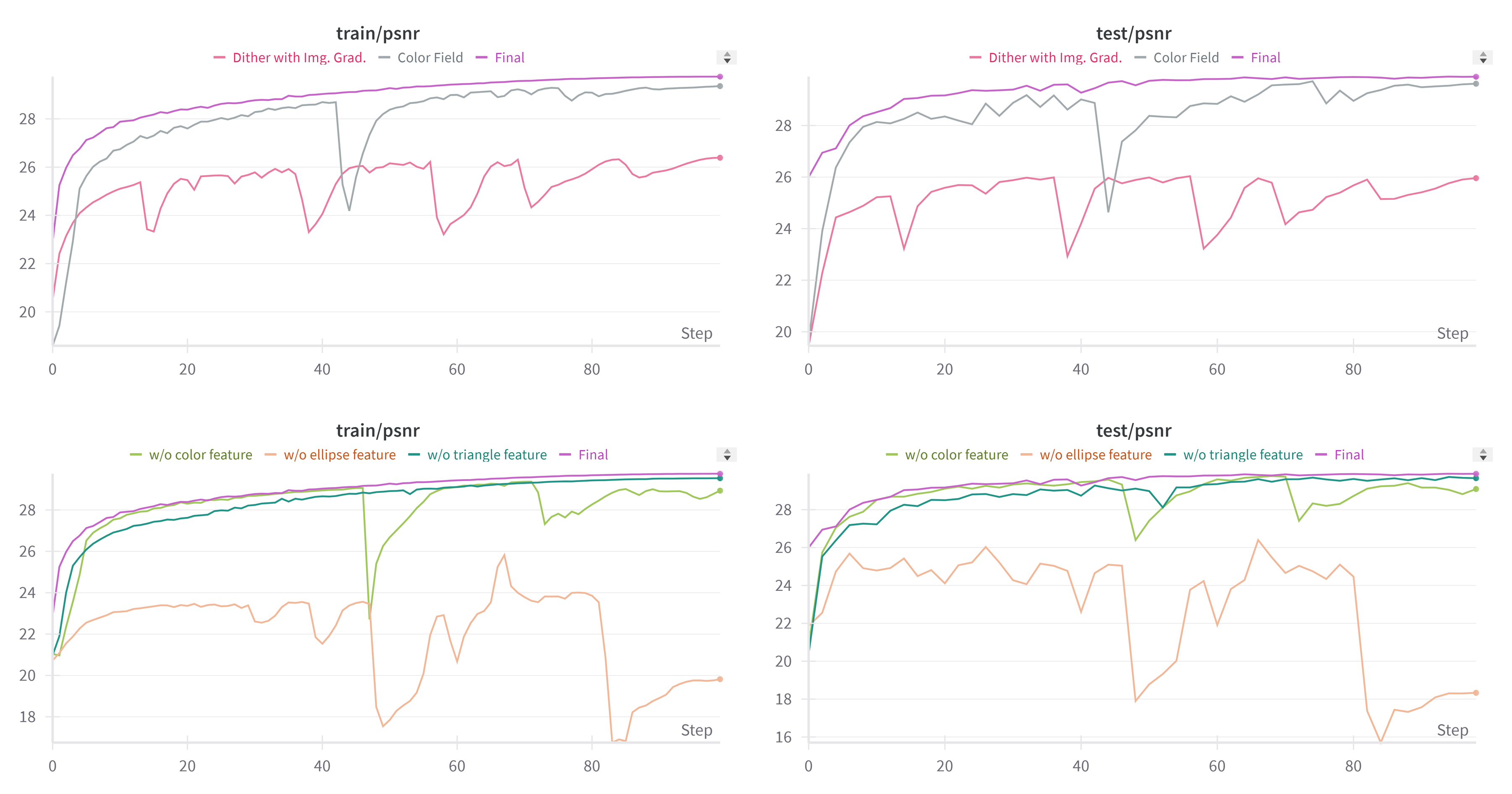}
\vspace{-1.0cm}
\end{center}
   \caption{PSNR curves during training process.
    }
\label{fig:ablation_line}
\end{figure*}

In this section, we provide additional experimental results to validate the settings discussed in the main paper.

\subsection{Additional Ablation Study}

In the main paper, we noted that while certain settings may not significantly impact final evaluation results, they can introduce instability during training, making the model prone to local optima or even training failure. In Fig.~\ref{fig:ablation_line}, we present PSNR curves for both the training and test sets to illustrate these effects.

The \textbf{top row} corresponds to the training curves for ``Dither with Img. Grad." and ``Color Field," as discussed in Tab.~3 of the main paper. Compared to our final model, predicting Gaussian colors directly instead of opacity leads to slower convergence and a tendency to get trapped in local optima during mid-stage training. While the final model performance remains similar, the opacity-based approach results in a more efficient training process. Furthermore, ``Dither with Img. Grad." struggles due to the significant gap between image gradients and the true Gaussian distribution density, leading to both lower performance and a higher likelihood of convergence issues.

The \textbf{bottom row} shows an ablation study on feature organization, evaluating the necessity of different features. We use image features as the default part and analyze the effects of including or excluding \textit{color features}, \textit{ellipse features}, and \textit{triangle features}. The results show that ellipse features are crucial for the final performance, while triangle and color features have a smaller impact but still contribute to faster convergence and improved training stability by preventing local optima.

\subsection{Additional Visualization Results}

Due to space limitations in the main paper, we provide additional rendering visualizations and comparisons here. We randomly select several images from both the Kodak and Div2k datasets. 

In Fig.~\ref{fig:more_vis}, we compare Gaussian representation initialization using our network (\textbf{top row}) and GaussianImage’s random initialization strategy (\textbf{bottom row}). The visualization includes the initial results, rendering outputs at the 2s mark (including initialization time), and detailed comparisons. From the visual results, our method produces high-quality initialization, leading to superior rendering even before fine-tuning. At the 2s mark, it achieves \textit{near-Ground-Truth} quality, demonstrating finer details compared to GaussianImage.

\begin{figure*}[ht!]
\begin{center}
\includegraphics[width=0.9\linewidth]{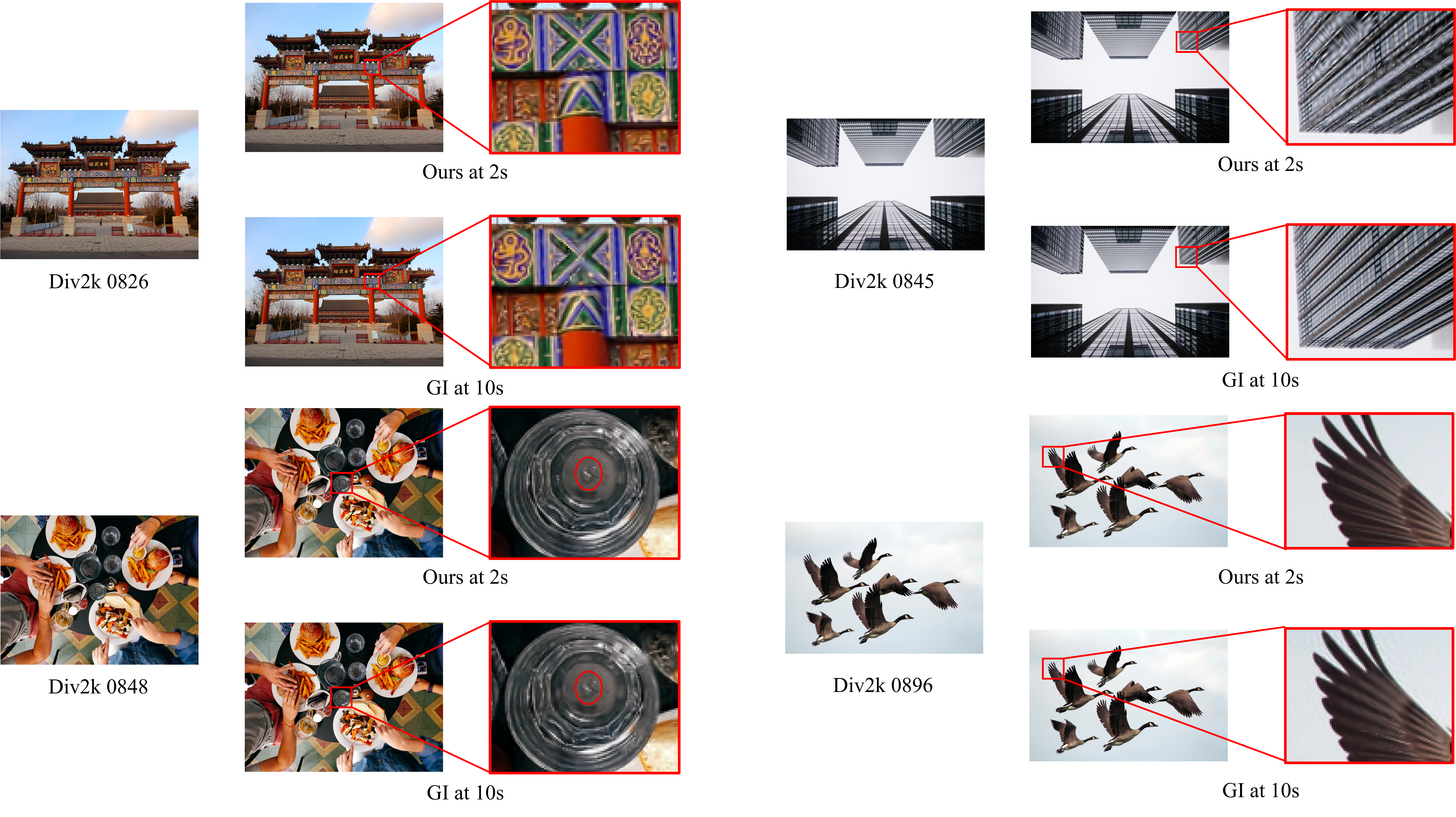}
\vspace{-0.5cm}
\end{center}
   \caption{ More visualization Results. Please zoom in for more details.
    }
\label{fig:more_vis_2}
\end{figure*}

In Fig.~\ref{fig:more_vis_2}, we compare the output of our network-based initialization at the 2s mark with the output of random initialization at 10s. From the detailed comparisons, it is evident that our method achieves results at 2s that nearly achieve or even surpass those obtained after 10s of training with random initialization, further demonstrating the efficiency of our approach.

\begin{figure*}[ht!]
\begin{center}
\includegraphics[width=0.9\linewidth]{pics/MoreVis.pdf}
\vspace{-0.5cm}
\end{center}
   \caption{ More visualization Results. Please zoom in for more details.
    }
\label{fig:more_vis}
\end{figure*}

{
    \small
    \bibliographystyle{ieeenat_fullname}
    \bibliography{main}
}

%% file: sec/0_abstract.tex
\begin{abstract}
Implicit Neural Representation (INR) has demonstrated remarkable advances in the field of image representation but demands substantial GPU resources. GaussianImage recently pioneered the use of Gaussian Splatting to mitigate this cost, however, the slow training process limits its practicality, and the fixed number of Gaussians per image limits its adaptability to varying information entropy. To address these issues, we propose in this paper a generalizable and self-adaptive image representation framework based on 2D Gaussian Splatting. Our method employs a network to quickly generate a coarse Gaussian representation, followed by minimal fine-tuning steps, achieving comparable rendering quality of GaussianImage while significantly reducing training time. Moreover, our approach dynamically adjusts the number of Gaussian points based on image complexity to further enhance flexibility and efficiency in practice. Experiments on DIV2K and Kodak datasets show that our method matches or exceeds GaussianImage’s rendering performance with far fewer iterations and shorter training times. Specifically, our method reduces the training time by up to one order of magnitude while achieving superior rendering performance with the same number of Gaussians. Code is availiable at \href{https://github.com/whoiszzj/Instant-GI}{https://github.com/whoiszzj/Instant-GI}
\end{abstract}

%% file: sec/1_intro.tex
\section{Introduction}
\label{sec:intro}

\begin{figure}[ht!]
\begin{center}
\includegraphics[width=0.95\linewidth]{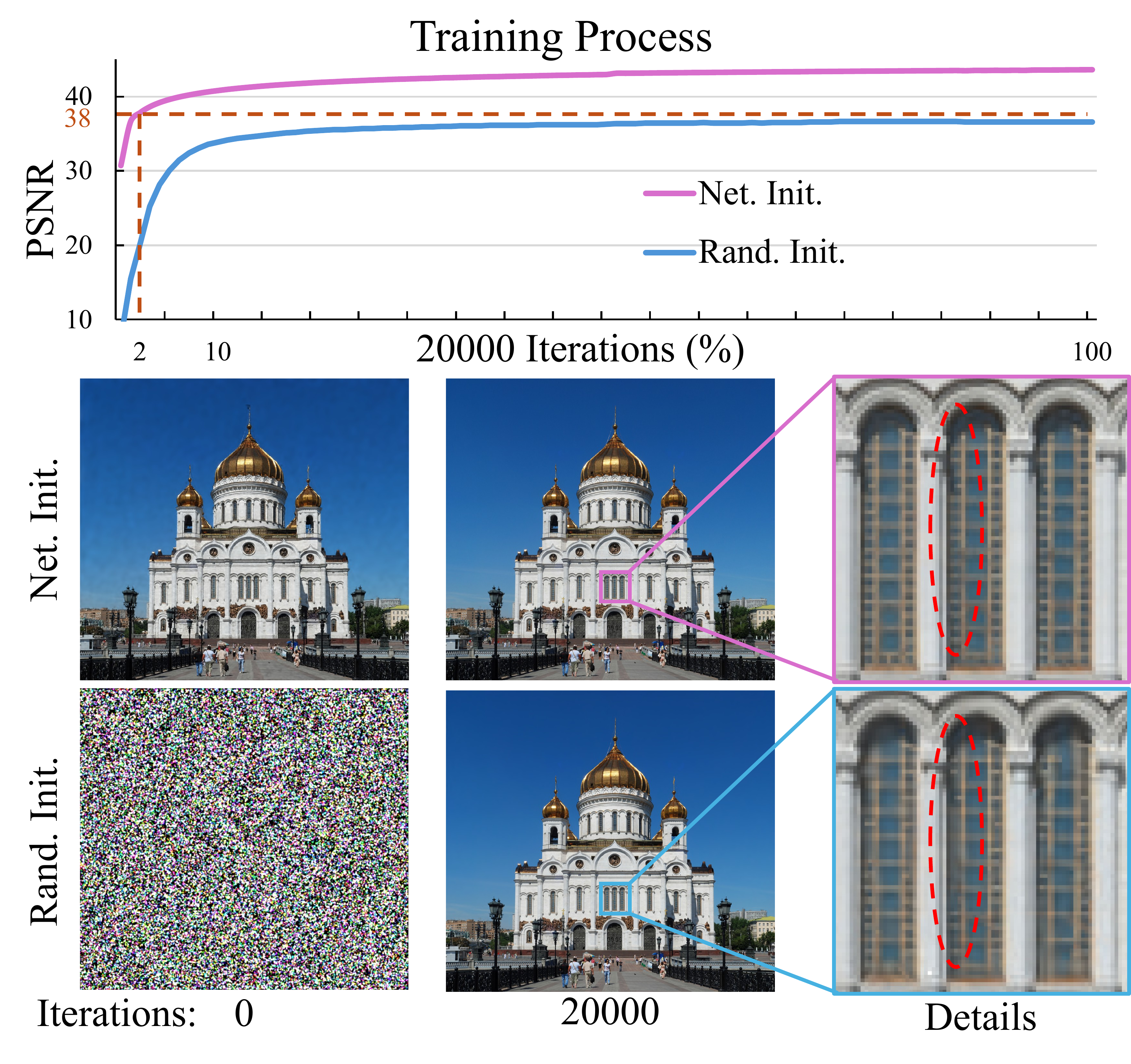}
\end{center}
 \vspace{-0.5cm}
   \caption{
   \textbf{Teaser.}
   Comparison of the network-based initialization (Net. Init.) and the random initialization (Rand. Init.) regarding training efficiency and rendering quality using the same number of Gaussians. The top graph shows that Net. Init. converges faster and achieves higher quality. The bottom row visualizes the initialization state (left), results after 20,000 iterations (middle), and zoomed-in details (right), highlighting the improved detail preservation and efficiency of our method.
   }
\label{fig:teaser}
\end{figure}

Image representation is an important task in computer vision, which has numerous applications in the fields such as image compression~\cite{takikawa2023compact, GaussianImage}, super-resolution~\cite{li2022adaptive, wu2023learning}, deblurring~\cite{Kong_2023_CVPR, mou2022deep}, and so on. The emergence of Implicit Neural Representation (INR) has shifted image representation from traditional methods like grid graphics to the more recent approach using multilayer perceptrons (MLPs). This method leverages the local continuity of the data to map the input coordinates to their corresponding output values.

Before the introduction of methods like GaussianImage (GI) \cite{GaussianImage}, INR techniques relied primarily on high-dimensional MLPs to capture fine image details. However, this strategy leads to slower training, higher memory consumption, and longer decoding times, making it less suitable for real-world applications. To address these limitations, GI draws inspiration from 3D Gaussian Splatting~\cite{3DGS} (3DGS), utilizing discrete Gaussian primitives for image information extraction and fitting. In addition, it employs an accumulated blending-based rasterization approach to enable fast rendering, facilitating rapid decoding. For the convenience of subsequent descriptions, we define ``Gaussian Decomposition'' as the process of obtaining a Gaussian representation from an image through certain operations to represent its content.

Although GI achieves decoding speeds exceeding 1000 FPS, it still faces several challenges. Firstly, the Gaussian primitives are initialized randomly, requiring a relatively long training time to achieve a stable fitting result. 
%As noted in the paper, the encoding is about 211 seconds. 
Secondly, the number of Gaussian primitives must be predetermined before training, which poses challenges for images with varying levels of information entropy. High-entropy images, which contain rich details and sharp transitions, require a larger number of Gaussians to capture fine details. Conversely, low-entropy images, characterized by smooth variations and fewer details, do not require as many Gaussians. Using too few Gaussians for high-entropy images may lead to loss of detail, while an excessive number for low-entropy images results in unnecessary storage overhead and reduced rendering efficiency.

To resolve these issues and improve the rendering performance of GI, we propose to utilize a generalizable network to initialize Gaussian representation, enabling a more efficient Gaussian Decomposition through a single feedforward pass. To this end, we propose a novel pipeline for Gaussian Decomposition, named as Instant GaussianImage (Instant-GI). Specifically, given an input image, we first employ a feature extraction network to generate a feature map, from which a Gaussian Position Probability Map (PPM) is derived. The PPM is then discretized using the Floyd–Steinberg dithering~\cite{DitheringGPU} algorithm, allowing for a flexible number of Gaussians to adapt to images with varying levels of information entropy. Finally, based on the dithering result and feature map, the full set of Gaussian attributes are predicted, and the entire network is trained via a differentiable rasterization for self-supervised learning.

With the proposed generalizable network, a high-quality Gaussian representation can be quickly initialized for an image. Followed by minimal fine-tuning steps, it can match or exceed the results obtained by GI through long training, as shown in \cref{fig:teaser}.  Furthermore, our method adaptively determines the number of Gaussians based on the information entropy of the image, ensuring both efficiency and flexibility between different images. Our main contributions are summarized as follows.
\begin{itemize} 
    \item We propose a generalizable network for fast Gaussian Decomposition, enabling efficient initialization that significantly reduces training time.
    \item We introduce the Floyd–Steinberg dithering algorithm to discretize the Gaussian PPM, allowing for adaptive and optimal spatial distribution of point cloud based on an image's information entropy.
    \item We develop a novel pipeline for Gaussian Decomposition that generates a high-quality Gaussian representation in a single feedforward pass.
    \item Numerical experiments on DIV2K and Kodak datasets demonstrate that our method, with much shorter training times, achieves results comparable or superior to GI.
\end{itemize}

%% file: sec/2_related.tex
\section{Related Work}
\label{sec:related_work}

%-------------------------------------------------------------------------
\subsection{Implicit Neural Representation}
Implicit Neural Representation  hass been widely applied in various domains due to their ability to represent continuous spatial information, including 3D scene representation~\cite{NeRF, barron2021mip, wang2021neus, barron2023zipnerf}, image and video representation~\cite{I-NGP, Wire, LIIF, chen2023hnerv, chen2021nerv}, compression~\cite{kuznetsov2021neumip, vaidyanathan2023random, takikawa2023compact}, and super-resolution~\cite{dong2015image, lim2017enhanced, hu2019meta, li2022adaptive}.  In image representation, \cite{klocek2019hypernetwork} introduces a hypernetwork-based functional representation, mapping coordinates to colors for continuous reconstruction. To mitigate spectral bias, \cite{tancik2020fourier} proposes Fourier feature mapping, while \cite{SIREN} leverages periodic activation functions for improved detail modeling. \cite{LIIF} bridges discrete and continuous representations by predicting RGB values from coordinates and local deep features.  Recent works have explored adaptive and efficient INR designs. \cite{martel2021acorn} proposes a hybrid implicit-explicit model with multiscale decomposition for gigapixel images. \cite{I-NGP} employs multiresolution hash encoding to accelerate training and rendering. \cite{Neurbf} introduces neural fields with adaptive radial bases, improving detail capture. Additionally, \cite{belhe2023discontinuity} encodes discontinuities using Bézier curves and triangular meshes to enhance sharp feature preservation and compression efficiency.

%-------------------------------------------------------------------------
\subsection{Gaussian Splatting}
3DGS~\cite{3DGS} has gained significant attention in 3D scene reconstruction due to its fast training, efficient rendering, and high-quality results. Its differentiable rasterization process has also been applied to various domains, including depth map rendering for geometric reconstruction~\cite{huang20242d, yu2024gaussian, chen2024pgsr, zhang2024rade}, 3D model generation via diffusion~\cite{DiffGS, li2023gaussiandiffusion, tang2023dreamgaussian}, and 3D scene editing~\cite{chen2024gaussianeditor, jaganathan2024ice, wang2024view}.  Several works have extended 3DGS to enhance its effectiveness. Scaffold-GS~\cite{Scaffold-GS} integrates anchor points and scene features to improve Gaussian property prediction. MVSGaussian~\cite{MVSGaussian} leverages MVSNet for image-based feature extraction, followed by 3D convolution to estimate spatial point positions and predict Gaussian attributes. DiffGS~\cite{DiffGS} introduces a disentangled representation of 3D Gaussians, enabling flexible and high-fidelity Gaussian generation for rendering.

%-------------------------------------------------------------------------
\subsection{GS-based Image Representation}
In recent years, several methods have integrated Gaussian Splatting with image representation. One of the most notable is GaussianImage~\cite{GaussianImage}, which replaces traditional sorting-based splatting with an accumulated blending-based rasterization method. This enables faster training and rendering while achieving high-ratio image compression.  Image-GS~\cite{ImageGS} adopts a similar approach, computing grayscale gradients to generate probability values, which are then used for CDF-based Gaussian sampling and initialization. GaussianSR~\cite{GaussianSR} uses Gaussian Splatting's interpolation capabilities for super-resolution by converting an image into a feature map, generating a high-resolution feature representation and reconstructing the final output with a decoder. Mirage~\cite{Mirage} extends GI by projecting 2D images into 3D space, employing flat-controlled Gaussians for precise 2D image editing.

In this paper, we adopt GI as the baseline of our proposed work. Since our Gaussian model remains consistent with GI, all compression techniques for GI from ~\cite{GaussianImage} can be applied directly to our method. Therefore, we do not elaborate on this aspect. Instead, our primary goal focuses on fast and effective initialization to significantly enhance training speed and rendering performance.

%% file: sec/3_method.tex
\section{Proposed Method}
\label{sec:methodology}

%------------------------------------------------------------------------

\subsection{Preliminaries}
\label{preliminaries}
GI refines the Gaussian model based on 3DGS to enhance image representation performance,  and  our proposed method is also built upon this foundation. Note that a 2D Gaussian is characterized by the following attributes: position \(\boldsymbol\mu\in \mathbb{R}^2\), 2D covariance matrix \( \boldsymbol\Sigma \in \mathbb{R}^{2 \times 2}\), color parameter \(\boldsymbol c \in \mathbb{R}^3\), and opacity \(o \in \mathbb{R}\).

For the positive semidefinite covariance matrix, GI offers two different parameterization approaches. The first approach employs Cholesky factorization, directly decomposing the covariance matrix into a lower triangular matrix \( \boldsymbol L \), such that \( \boldsymbol\Sigma = \boldsymbol L \boldsymbol L^T \). The second approach, similar to 3DGS, decomposes \( \boldsymbol \Sigma \) into a scaling matrix \( \boldsymbol S \) and a rotation matrix \( \boldsymbol R \): 
\begin{equation}
    \boldsymbol \Sigma = (\boldsymbol R \boldsymbol S)(\boldsymbol R \boldsymbol S)^T,
\end{equation}
where
\begin{equation}
\mathbf{R} =
\begin{bmatrix}
\cos(\theta) & -\sin(\theta) \\
\sin(\theta) & \cos(\theta)
\end{bmatrix}
, \quad
\mathbf{S} =
\begin{bmatrix}
s_1 & 0 \\
0 & s_2
\end{bmatrix}
.
\label{eq:rotation_scaling}
\end{equation}
For ease of transformation, our method primarily adopts the \( \boldsymbol{RS} \) parameterization for learning and representation.

The learnable Gaussian parameters are optimized through differentiable rasterization. Unlike 3DGS, which needs to handle occlusion, GI does not require opacity accumulation or Gaussian sorting. Consequently, the rendering equation is given by:
\begin{equation}
\boldsymbol{C}_i = \sum_{n \in \mathcal{N}} \mathbf{c}_n \cdot \alpha_n
= \sum_{n \in \mathcal{N}} \mathbf{c}_n \cdot o_n \cdot \exp(-\sigma_n),
\end{equation}
where \( \mathcal{N} \) represents the set of Gaussians contributing to the rendering of the pixel, and
\begin{equation}
\alpha_n = o_n \cdot \exp(-\sigma_n), \quad
\sigma_n = \frac{1}{2} \mathbf{d}_n^{T} \Sigma^{-1} \mathbf{d}_n,
\end{equation}
where \( \mathbf{d}_n \in \mathbb{R}^2 \) denotes the displacement between the pixel center and the projected 2D Gaussian center.
Since both opacity and color are learnable parameters, GI integrates them together, which can be expressed as:
\begin{equation}
\boldsymbol{C}_i = \sum_{n \in \mathcal{N}} \mathbf{c}_n' \cdot \exp(-\sigma_n).
\end{equation}

In this case, the parameters of the Gaussian model consist of the position, covariance matrix, and color attributes -- a total of eight learnable parameters. During initialization, the number of Gaussians \( N \) is determined and remains fixed throughout training.  The position parameters \(\boldsymbol\mu\) are uniformly sampled from \([-1,1]\). The scaling parameters \(\boldsymbol s =(s_1,s_2)\) are initialized randomly in \([0.5, 1.5]\), while the rotation angle \(\theta\) is drawn from \([0,1]\). Similarly, the color attributes \(\boldsymbol c \) are randomly initialized within \([0,1]\).  Such purely random initialization lacks prior guidance, making the optimization process slower in deriving an accurate Gaussian representation of the image.

\begin{figure*}[ht!]
\begin{center}
\includegraphics[width=0.9\linewidth]{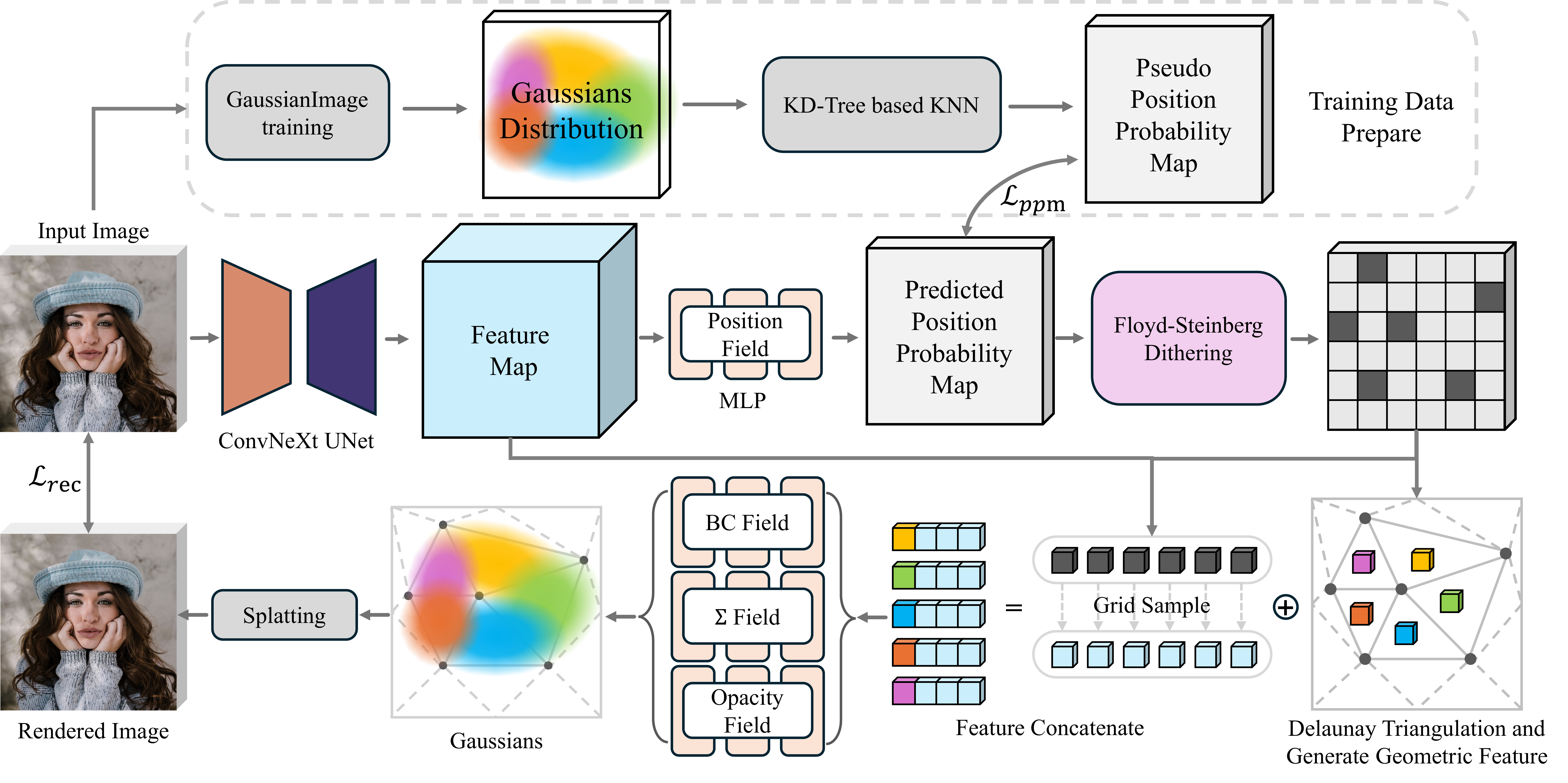}
\vspace{-0.5cm}
\end{center}
   \caption{\textbf{Overview of our Instant-GI pipeline.} Given an input image, we first extract a feature map using a ConvNeXt-based UNet. The Position Probability Map (PPM) is then predicted by an MLP-based Position Field and discretized via Floyd–Steinberg Dithering. The resulting points are structured using Delaunay Triangulation, from which geometric features are extracted. Gaussian attributes, including position, scaling, rotation, and opacity, are predicted through dedicated MLP fields (BC Field, $\Sigma$ Field, and Opacity Field). The final Gaussians are rendered via splatting, supervised by the RGB loss $\mathcal{L}_{rec}$ to ensure reconstruction quality. Additionally, a pre-trained Gaussian representation from GI is used to generate a Pseudo PPM, which supervises position probability learning via $\mathcal{L}_{ppm}$.
}
\label{fig:pipeline}
\end{figure*}

\subsection{Overview}
Our Instant-GI framework is illustrated in Fig.~\ref{fig:pipeline}. In the following sections, we introduce the main modules of our framework. Given an image of size \( W \times H \), our objective is to obtain its Gaussian representation using Instant-GI. We begin by employing a ConvNeXt-based~\cite{liu2022convnet} UNet to extract features from the image. Based on these extracted features, a self-adaptive number of Gaussians are generated. Then, the corresponding parameters are produced for each Gaussian to ensure that the splatting process generates a rendering result consistent with the input image. To achieve this goal, our work broadly consists of three parts: (1) how a self-adaptive number of Gaussians is generated from the extracted features, (2) how the features of each Gaussian are structured and transformed into appropriate Gaussian parameters, and (3) which loss functions are used to ensure that the entire network is trained effectively.

\subsection{Self-Adaptive Gaussian Samples}
\label{sec:self_adaptive_gaussians_sample}
To enable the algorithm to adaptively determine the number of Gaussians for images with varying levels of information entropy, ensuring consistent rendering performance across different images, intuitively:
\begin{itemize}
    \item The higher the information entropy of an image, the greater the number of Gaussians required.
    \item Within different regions of an image, areas with higher entropy should have a higher density of Gaussian.
\end{itemize}

A natural approach is to compute the color gradient of the image and use it to determine the Gaussian density, ensuring that regions with higher color gradients contain more Gaussians to capture high-frequency information. For instance, \cite{ImageGS} computes the probability of generating a Gaussian based on the gradient of the pixel and sampled with Cumulative Distribution Function (CDF).
% :
% \begin{equation}
% \mathbb{P}_{\text{init}}(\mathbf{x}) =
% \frac{(1 - \lambda_{\text{init}}) \cdot \|\nabla I(\mathbf{x})\|_2}
% {\sum_{h=1}^{H} \sum_{w=1}^{W} \|\nabla I(\mathbf{x}_{h,w})\|_2}
% + \frac{\lambda_{\text{init}}}{H \cdot W},
% \end{equation}
% where $\lambda_{\text{init}} \in [0, 1] $ balances local content adaptivity and uniform coverage.
However, the distribution of an image's Gaussian representation does not strictly correspond to variations in the color gradient. While the color gradient may exhibit abrupt changes, the variation in Gaussian density tends to be smoother (demonstrated in \cref{fig:probability_map} in \cref{sec:experiments}). As a result, directly deriving the probability of Gaussian generation from the gradient is suboptimal. To address this, we adopt a data-driven approach by training a network to predict the probability of Gaussian generation more effectively.
%Therefore, we aim to directly utilize a pre-trained Gaussian representation to generate a pseudo Position Probability Map (PPM) that indicates the distribution of Gaussians. 

\noindent\textbf{Obtain Pseudo PPM.}
\label{sec:obtain_pseudo_ppm}
To train a network capable of predicting a Position Probability Map (PPM) that indicates the spatial distribution of Gaussians, we first generate high-quality training data by Gaussian Decomposition. After that, we compute the Gaussian density at each pixel position \(\mathbf{x}\) in the image. Specifically, for each \(\mathbf{x}\), we estimate the minimal radius of a circle centered on \(\mathbf{x}\) that encompasses \( K \) Gaussians, then the local Gaussian density $D_{\mathbf{x}}$ is defined as the unit density of Gaussians in this circle:
\begin{equation}
    D_{\mathbf{x}} =  \frac{K}{\pi \cdot {\max\limits_{i} \left( \|\mathbf{p}_{\mathbf{x}} - \mathbf{p}_i\| \right) }^2},
\end{equation}
where \(\mathbf{p}\) represents the position coordinates, and \( i \) indexes the top-\( K \) nearest neighbors of \(\mathbf{x}\). In our experiments, we set \( K = 10 \).  
Next, we convert this density map into a PPM, which determines the probability of generating a Gaussian at each pixel location:
\begin{equation}
    \mathbb{P}_{\text{pseudo}}(\mathbf{x}) \propto D_{\mathbf{x}}.
\end{equation}
Finally, under the supervision of \( \mathbb{P}_{\text{pseudo}} \), we use an MLP network (Position Field in Fig.~\ref{fig:pipeline}) to predict \(\mathbb{P}_{\text{pred}}\). For more details of the generation of \( \mathbb{P}_{\text{pseudo}} \), please refer to the supplementary material.

\noindent \textbf{Dither with} \(\mathbb{P}_{\text{pred}}\)\textbf{.}
\label{sec:dithering}
The key challenge is how to sample Gaussians based on the PPM. A straightforward approach is thresholding—sampling a Gaussian if its probability exceeds a certain value. However, this ignores spatial dependencies, leading to excessive Gaussians in high-entropy regions and sparse coverage in low-entropy regions, resulting in underfitting and overfitting~\cite{3DGS}. An alternative is CDF sampling with a fixed number of Gaussians, as in \cite{ImageGS}. While it works when high- and low-entropy regions are balanced, it degrades rendering quality when high-entropy regions dominate, reducing point density where detail is needed. Conversely, in low-entropy images, excessive points are allocated to smooth regions, leading to redundancy (refer to supplementary material for details). 

To address this, before generating the Gaussian, we adopt the Floyd–Steinberg Dithering algorithm~\cite{DitheringGPU} to adaptively discretize the PPM. The map is firstly divided into \( k \times k \) patches, with each patch assigned its highest probability as the representative value. We then apply dithering and use the centers of the activated patches as sampling points, ensuring a balanced spatial distribution.

\begin{figure}[t!]
\begin{center}
\includegraphics[width=0.95 \linewidth]{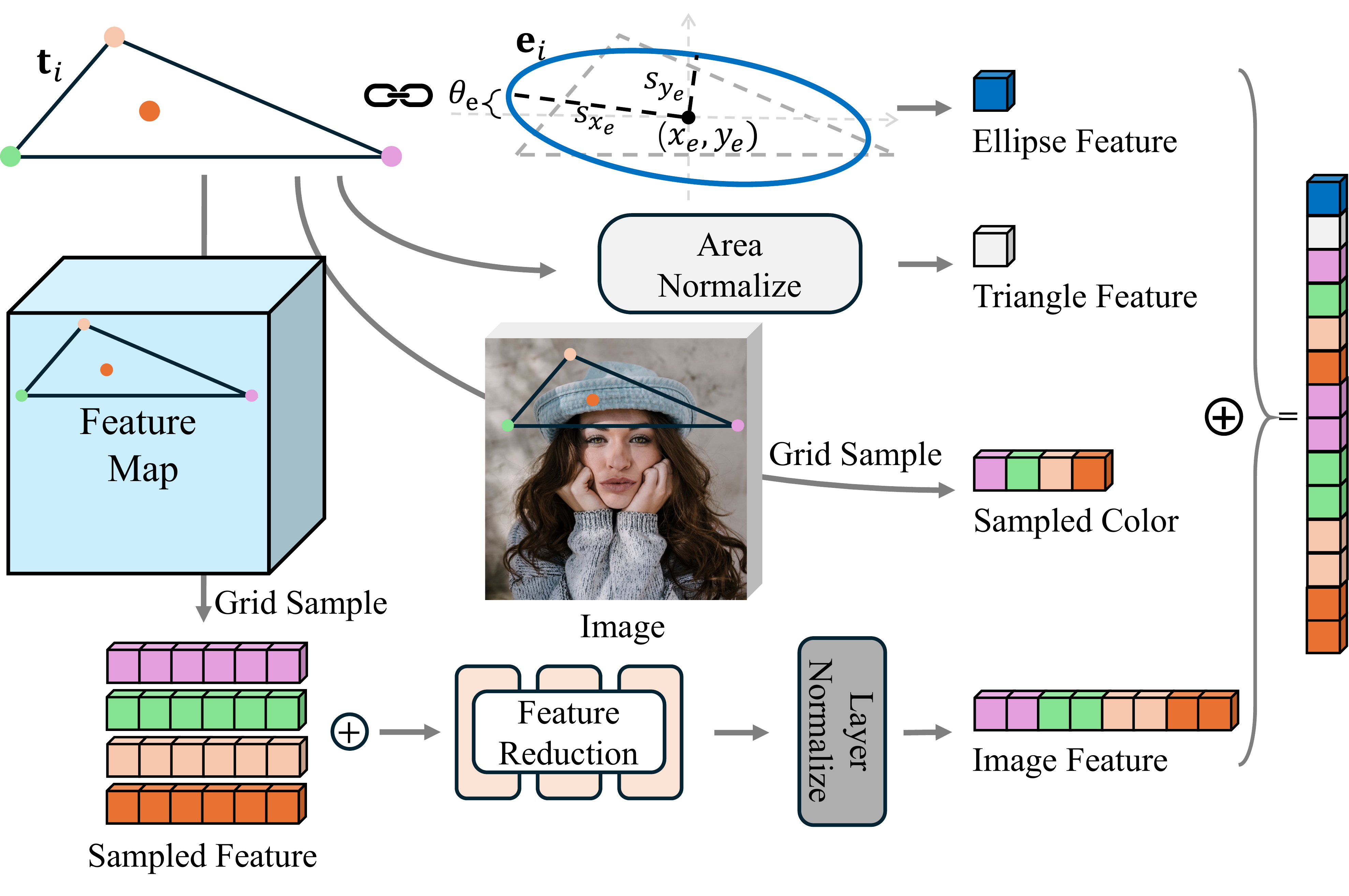}
\end{center}
\vspace{-0.5cm}
   \caption{
   \textbf{Feature organization for MLP input.} We construct the MLP input by combining multiple feature components. Geometric features are extracted from the triangle \( \mathbf t_i \) and its fitted ellipse \( \mathbf e_i \), while grid sampling provides local image features and sampled colors from the input image. The image features undergo feature reduction and layer normalization before concatenation.
    }
\label{fig:feature_organization}
\end{figure}

\subsection{Transformation from Feature to Gaussians}
\label{sec:transformation_from_feature_to_gaussians}
\noindent \textbf{Ellipse Fitting.}
\label{sec:ellipse_fitting}
From the previous steps, we obtain a set of discrete points whose distribution reflects the image's entropy information. However, to fully define the Gaussians, we still need to predict their scaling, rotation, and color. 

Since the scaling is measured in pixel units, directly predicting an absolute value with the network is hard work. Instead, a reference value is required to constrain the learning range, similar to \cite{Scaffold-GS}. To better determine the learning range for scaling, we introduce the Delaunay Triangulation~\cite{DelaunayTri} method for help based on two insights:
\begin{itemize}
    \item Given that the scaling defines a Gaussian as an ellipse, the overlap between ellipses should be minimized to reduce the network's learning complexity.
    \item Simultaneously, all ellipses should fully cover the entire image to prevent underfitting.
\end{itemize}
Specifically, we perform Delaunay Triangulation on the discrete points, using the resulting triangles as fundamental feature processing primitives as they can guarantee full coverage without overlap. 
%For each triangle, we fit an ellipse~\cite{fitEllipse} to approximate its overall shape. 
%This fitted ellipse provides an initial estimate for Gaussian parameters, primarily supplying baseline scaling and rotation information. 
For the obtained set of triangle representations \( \{\mathbf{t}_i\}_{i=0}^{T} \), where each \( \mathbf{t}_i \) consists of the three vertex coordinates of the corresponding triangle:
\begin{equation}
    \mathbf{t}_i = \begin{bmatrix} x_a & y_a \\ x_b & y_b \\ x_c & y_c \end{bmatrix}.
\end{equation}
Next, we compute the midpoints of the three edges of each triangle to extract additional geometric information. Using these six points, we perform ellipse fitting~\cite{fitEllipse} to obtain the set of ellipses \( \{\mathbf{e}_i\}_{i=0}^{T} \), where each ellipse is characterized by its position, major and minor axes, and rotation parameters:
\begin{equation}
    \mathbf{e}_i = \{(x_e, y_e), (s_{x_e}, s_{y_e}), \theta_{e} \}.
\end{equation}
For more details, please refer to the supplementary material.

\noindent \textbf{Feature Organization}
\label{sec:feature_organization}
After obtaining the fitted ellipse for each triangle, the next step is to extract the corresponding feature information for each primitive. 
%Specifically, each primitive's features are derived from four components, as illustrated in Fig.~\ref{fig:feature_organization}. 
As illustrated in Fig.~\ref{fig:feature_organization}, given the triangle \( \mathbf{t}_i \) and its fitted ellipse \( \mathbf{e}_i \), we compute the triangle and ellipse features, which serve as geometric attribute inputs to the network. Additionally, we sample both the image and the feature map at the three triangle vertices and the center point to obtain the corresponding color and feature values. To prevent an excessively high feature dimension, we apply feature reduction and layer normalization to the sampled features. The four feature components are concatenated to form the final feature vector, which is fed into the Gaussian fields. For more details, please refer to the supplementary material.

\noindent \textbf{Gaussian Fields.}
\label{sec:gaussian_fields}
After extracting the features of each primitive, we predict all Gaussian attributes based on these features, including position, scaling, rotation, and color.

When predicting position information, we use the ``BC Field'' in Fig.~\ref{fig:pipeline} to compute the barycentric coordinates of the Gaussian center \( \mathbf{bc} \in \mathbb{R}^3 \). The final Gaussian position \( \boldsymbol\mu_i \) is obtained as:
\begin{equation}
    \boldsymbol\mu_i = \mathbf{bc}_i \cdot \mathbf{t}_i.
\end{equation}

When predicting the scaling and rotation parameters, we use a single MLP, denoted as the ``\(\Sigma\) Field'' in Fig.~\ref{fig:pipeline}. Given the major and minor axes \((s_{x_e}, s_{y_e})\) and the rotation angle \(\theta_e\) of a fitted ellipse, our predictions include the scaling adjustment factors \((O_{s_x}, O_{s_y})\) and the rotation bias \( O_\theta \). The activation function applied to \((O_{s_x}, O_{s_y})\) is defined as:
\begin{equation}
f_s(x) =
\begin{cases}
0.5 \cdot e^{0.5x}, & x < 0, \\
-\frac{1}{2} \log(1 + e^{-0.5x + 5}) + 3, & x \geq 0.
\end{cases}
\end{equation}
The final scaling \( \boldsymbol{s}_i = (s_x, s_y) \) is computed as:
\begin{equation}
    s_x = f_s(O_{s_x}) \cdot s_{x_e}, \quad s_y = f_s(O_{s_y}) \cdot s_{y_e}.
\end{equation}
The final rotation \( \theta_i \) is obtained as:
\begin{equation}
\begin{aligned}
\theta_i &= \sigma \left( \sigma^{-1}(e_{\theta}) + \tanh(O_\theta) \right), \\
\end{aligned}
\end{equation}
where $\sigma$ is the sigmoid function, and $\sigma^{-1}$ is the inverse.

When predicting color attributes, we found that directly regressing RGB values from features results in slow convergence during early training stages and may even lead to training failure. Thus, we predict the opacity attribute \( o \) with the ``Opacity Field'' in Fig.~\ref{fig:pipeline} instead of directly predicting color. The final color attribute \( \boldsymbol{c}_i \) is then obtained by multiplying the color sampled from the image $I$ at the Gaussian position \( \boldsymbol{p}_i \) with the opacity:
\begin{equation}
    \boldsymbol{c}_i = I(\boldsymbol{p}_i) \cdot \sigma(o).
\end{equation}
At this point, all the attributes of each Gaussian primitive $\mathbf{G}_i$ have been initialized as:
\begin{equation}
    \mathbf{G}_i = \{\boldsymbol{p}_i, \boldsymbol{s}_i, \theta_i, \boldsymbol{c}_i\}.
\end{equation}

\subsection{Loss Function}
\label{sec:loss_function}
To ensure network training, the initialized Gaussian primitives are rendered into an image using the splatting technique. The primary objective of the network is image reconstruction, optimized via the loss function \( \mathcal{L}_{rec} \), which combines the \( \mathcal{L}_2 \) loss and the D-SSIM term:
\begin{equation}
    \mathcal{L}_{rec} = \lambda \cdot \mathcal{L}_2 + (1 - \lambda) \cdot \mathcal{L}_{D-SSIM}.
\end{equation}
Furthermore, to supervise the PPM, we introduce the loss function \( \mathcal{L}_{ppm} \), formulated as a Focal MSE Loss:
\begin{equation}
    \mathcal{L}_{ppm} = \alpha \cdot \left(1 - e^{-|\mathbb{P}_{\text{pseudo}} - \mathbb{P}_{\text{pred}}|}\right)^\gamma \cdot (\mathbb{P}_{\text{pseudo}} - \mathbb{P}_{\text{pred}})^2.
\end{equation}
The overall optimization objective is defined as:
\begin{equation}
    \mathcal{L} = \mathcal{L}_{rec} + \mathcal{L}_{ppm}.
\end{equation}

%% file: sec/4_experiment.tex
\section{Experiments}
\label{sec:experiments}

\subsection{Experimental Setup}
\noindent \textbf{Dataset}. Following GI, we evaluate our method on the Kodak~\cite{Kodak} and DIV2K~\cite{Div2k} datasets. The Kodak dataset consists of 24 images, each with a resolution of \( 768 \times 512 \). The DIV2K dataset contains a training set of 800 images and a test set of 100 images.  For network training, we utilize images from the DIV2K training set, applying bicubic downscaling at scales \( \times2 \), \( \times3 \), and \( \times4 \). During evaluation, to ensure consistency with GI and other methods, we conduct testing at a scale of \( \times2 \) on the test set, with image dimensions ranging from \( 408 \times 1020 \) to \( 1020 \times 1020 \).

\noindent \textbf{Implementation}.
%we adopt a UNet architecture based on ConvNeXt for feature extraction. The encoder employs the ConvNeXt base model, while the decoder outputs features with a final dimensionality of 64. The Position Field, BC Field, \( \Sigma \) Field, and Opacity Field are implemented as MLPs. Further details can be found in the supplementary material. 
%Once the PPM is obtained, to control the number of points at different levels, we divide the image into different minimum patch sizes (\( k \times k \)), enabling multi-level outputs. We set \( k=3 \) during training, while during testing, different levels such as \( k=3, 4, 5 \) can be used. To achieve this, max pooling is applied to downsample the PPM, followed by dithering using the Floyd–Steinberg Dithering algorithm. 
%To accelerate dithering, we implement a GPU-accelerated version based on~\cite{DitheringGPU}. After obtaining the sampled points via dithering, upsampling is applied to recover their positions at the original resolution. For loss parameter settings, we set \( \lambda=2 \), \( \alpha=1.0 \), and \( \gamma=2.0 \). We use the Adam~\cite{Adam} optimizer with a cosine annealing scheduler~\cite{loshchilov2016sgdr}, where the learning rate decays from \( 1 \times 10^{-3} \) to \( 1 \times 10^{-5} \) over 100 epochs. For the subsequent fine-tuning stage, we follow the same experimental settings as GI. Our algorithm is trained and tested on one NVIDIA A100 (40GB).
We set patch size \( k=3 \) in \cref{sec:self_adaptive_gaussians_sample} during training, while during testing, different levels such as \( k=3, 4, 5 \) can be used to further control the number of Gaussians. To accelerate Floyd–Steinberg dithering, we implement a GPU-accelerated version based on~\cite{DitheringGPU}.  For loss parameter settings, we set \( \lambda=2 \), \( \alpha=1.0 \), and \( \gamma=2.0 \). We use the Adam~\cite{Adam} optimizer with a cosine annealing scheduler~\cite{loshchilov2016sgdr}, where the learning rate decays from \( 1 \times 10^{-3} \) to \( 1 \times 10^{-5} \) over 100 epochs. For the subsequent fine-tuning stage, we follow the same experimental settings as GI. Our method is trained and tested on one NVIDIA A100 (40GB).
For more details, please refer to the supplementary material.

\begin{table}[ht!]
\centering
\resizebox{0.95\linewidth}{!}{
\setlength{\tabcolsep}{2pt}
\begin{tabular}{l|ccc|ccc}
\hline
Datasets       & \multicolumn{3}{c|}{Kodak}                                 & \multicolumn{3}{c}{DIV2K $\times$2}                               \\ \hline
Methods        & PSNR$\uparrow$ & MS-SSIM$\uparrow$ & Params(K)$\downarrow$ & PSNR$\uparrow$ & MS-SSIM$\uparrow$ & Params(K)$\downarrow$ \\ \hline
WIRE           & 41.47          & 0.9939            & 136.74                & 35.64          & 0.9511            & 136.74                \\
SIREN          & 40.83          & 0.9960            & 272.70                & 39.08          & 0.9958            & 483.60                \\
I-NGP          & 43.88          & 0.9976            & 300.09                & 37.06          & 0.9950            & 525.40                \\
NeuRBF         & 43.78          & 0.9964            & 337.29                & 38.60          & 0.9913            & 383.65                \\
3DGS           & 43.69          & 0.9991            & 3540.00               & 39.36          & 0.9979            & 4130.00               \\
GI  & 44.08          & 0.9985            & 560.00                & 39.53          & 0.9975            & 560.00                \\
GI$^\dag$ & 41.44          & 0.9979            & 342.86                & 40.26          & 0.9980            & 615.05                \\
Ours           & 42.92          & 0.9972            & 342.86                & 42.80          & 0.9982            & 615.05                \\ \hline
\end{tabular}
}
\caption{\textbf{Quantitative comparison results.} PSNR, MS-SSIM, and parameter count comparison across methods on Kodak and DIV2K\( \times 2 \). GI$^\dag$ means using the same number of Gaussians as our method but initializes parameters with GI’s approach.
}
\label{tab:kodak_div2k}
\end{table}

\begin{table*}[ht!]
\centering
\resizebox{0.7\linewidth}{!}{
\begin{tabular}{c|c|c|ccc|c|c|c}
\hline
\multirow{2}{*}{Image ID}      & \multirow{2}{*}{Gaussian Num.} & \multirow{2}{*}{Init. Method} & \multicolumn{3}{c|}{PSNR}           & \multirow{2}{*}{GPU Mem. (MB)} & \multirow{2}{*}{Training FPS} & \multirow{2}{*}{Testing FPS} \\ \cline{4-6}
                               &                                &                               & \multicolumn{1}{c|}{2s}    & \multicolumn{1}{c|}{10s}   & \multicolumn{1}{c|}{20s}   &                                &                               &                              \\ \hline
\multirow{2}{*}{0844$\times$2} & \multirow{2}{*}{60170}         & Rand. Init                  & \multicolumn{1}{c|}{35.14} & \multicolumn{1}{c|}{44.75} & \multicolumn{1}{c|}{45.90} & 410                            & 717                           & 3015                         \\ \cline{3-9} 
                               &                                & Net. Init.                     & \multicolumn{1}{c|}{46.68} & \multicolumn{1}{c|}{49.05} & \multicolumn{1}{c|}{49.58} & 3038 + 408                       & 447                           & 1969                         \\ \hline
\multirow{2}{*}{0879$\times$2} & \multirow{2}{*}{77690}         & Rand. Init                  & \multicolumn{1}{c|}{31.96}  & \multicolumn{1}{c|}{36.04} & \multicolumn{1}{c|}{36.47} & 454                            & 582                           & 2506                         \\ \cline{3-9} 
                               &                                & Net. Init.                     & \multicolumn{1}{c|}{37.41} & \multicolumn{1}{c|}{41.51} & \multicolumn{1}{c|}{42.49} & 3624 + 466                       & 344                           & 1536                         \\ \hline
\multirow{2}{*}{0858$\times$2} & \multirow{2}{*}{82766}         & Rand. Init                  & \multicolumn{1}{c|}{32.25} & \multicolumn{1}{c|}{36.58} & \multicolumn{1}{c|}{37.12} & 414                            & 736                           & 2733                         \\ \cline{3-9} 
                               &                                & Net. Init.                     & \multicolumn{1}{c|}{36.79} & \multicolumn{1}{c|}{39.72} & \multicolumn{1}{c|}{40.51} & 3038 + 414                       & 718                           & 2740                         \\ \hline
\end{tabular}
}
\caption{\textbf{Performance comparison results.} Comparison of different initialization methods (Rand. Init vs. Net. Init.) across multiple images in terms of PSNR at different time intervals (2s, 10s, 20s), GPU memory consumption, and FPS during training and testing.
}
\label{tab:performance}
\end{table*}

\begin{figure*}[ht!]
\begin{center}
\includegraphics[width=0.85\linewidth]{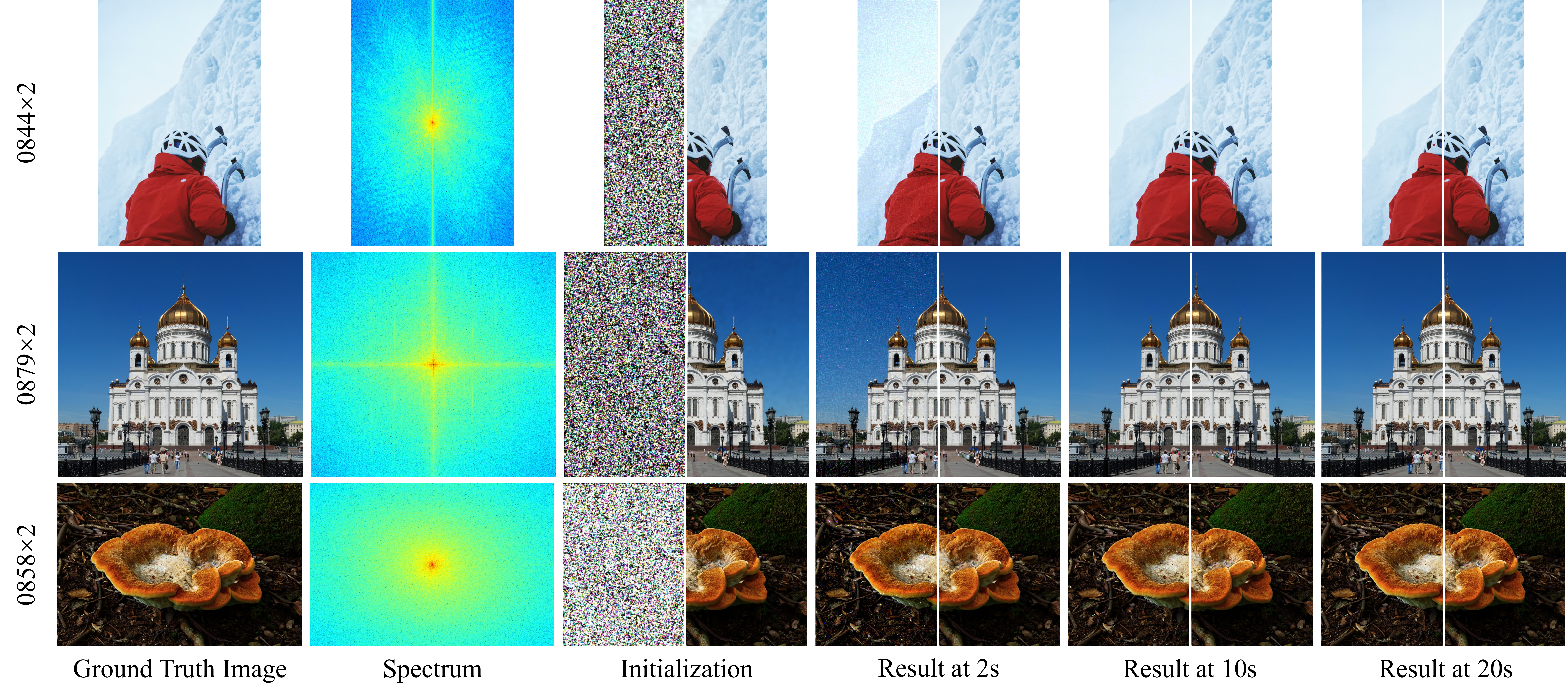}
\end{center}
\vspace{-0.5cm}
   \caption{\textbf{Reconstruction Progress Visualization.} Each row sequentially shows the GT image, frequency spectrum, initialization results, and results at 2s, 10s, and 20s. The results compare Rand. Init (left) and Net. Init. (right).
}
\label{fig:performance}
\end{figure*}

\subsection{Quantitative Results}
This experiment aims to demonstrate the upper bounder of rendering performance achievable by our method. We compare our approach with several recent image representation methods, with the specific results shown in \cref{tab:kodak_div2k}.  In this experiment, both our method and GI undergo extensive training (50,000 iterations). 
%Specifically, after initialization, we perform 50,000 iterations of optimization to obtain the final rendering results. 
% GI* denotes a variant where it only uses the same number of Gaussians as ours does. 
%we use our method to initialize the number of points for each image but apply GI's initialization method for optimization.

Firstly, our method achieves state-of-the-art (SOTA) performance on the DIV2K \( \times 2 \) dataset. Secondly, compared to GI$^\dag$, our method outperforms in final rendering quality under the same number of Gaussians. This is due to our initialization aligning better with image information entropy, reducing optimization complexity, and enhancing rendering quality. Thirdly, while our method does not surpass GI on the Kodak dataset, this is primarily due to Kodak’s lower resolution, leading to a 61\% reduction in Gaussian points. However, across both datasets, our method demonstrates a more stable performance (PSNR: 42.92 on Kodak, 42.80 on DIV2K \( \times 2 \)), proving its ability to dynamically adjust Gaussian counts based on information entropy for consistent image representation. In summary, our method enables a well-structured distribution with an optimized number of Gaussians, reducing the computational complexity of optimization while improving rendering quality.

\subsection{Performance \& Qualitative Results}
One of the key contributions of our algorithm is its ability to achieve usable rendering results within a short time. Therefore, we specifically analyze this performance in this subsection. In this experiment, we select three representative images: one with predominantly low-entropy information, one with a balance of high and low entropy, and one with predominantly high-entropy information. We compare GI with random initialization and our network-based initialization, reporting PSNR at 2s, 10s, and 20s (including initialization time) for both methods.  As shown in \cref{tab:performance}, our initialization significantly accelerates rendering. For images 0844\(\times\)2 and 0879\(\times\)2, the 2s rendering surpasses random initialization at 20s, achieving over a 10\(\times\) speed-up. Even for 0858\(\times\)2, which predominantly contains high-entropy regions and exhibits a spatial distribution similar to random initialization, our method still achieves a 5\(\times\) speed-up by predicting more accurate scaling, rotation, and color parameters. Regarding memory usage, our method requires approximately 3GB due to additional network computations, but this is negligible for modern GPUs.

In \cref{fig:performance}, we present the frequency spectrum analysis, initialization results, and rendered outputs at different time intervals for three selected images. From the initial rendering, our method already achieves high-quality results, with only minor deficiencies in fine details. By 2s, the rendering output becomes visually indistinguishable from the ground truth, whereas the random initialization method remains unconverged, exhibiting significant artifacts.  At 10s and 20s, both methods produce nearly identical results. However, upon closer inspection, fine-grained differences persist, as illustrated in \cref{fig:teaser}. Additional results can be found in the supplementary material.

Based on the above observations, we conclude that our algorithm achieves high-quality rendering immediately after network-based initialization. With only 2 seconds of initialization and fine-tuning, our method attains rendering quality comparable to GI after 20+ seconds of training, significantly improving the efficiency. %  of Gaussian decomposition.

\subsection{Ablation Study}

We conduct four ablation experiments to validate the rationality of our proposed method, as summarized in \cref{tab:ablation}. These experiments aim to answer the following questions:
\begin{itemize}
    \item Why can't the PPM be directly derived from image gradient but instead needs to be learned through network?  
    \item Why not directly predict color but learn opacity?  
    \item Why don't we initialize directly with the fitted ellipse?  
    \item If the number of points in an image is self-adaptive, how can different compression ratios be achieved?
\end{itemize}

\begin{table}[t!]
\resizebox{0.95\linewidth}{!}{
\setlength{\tabcolsep}{2pt}
\begin{tabular}{l|c|cc|cc}
\hline
\multirow{2}{*}{Method Describe} & \multirow{2}{*}{Gaussian Num.} & \multicolumn{2}{c|}{Initialization} & \multicolumn{2}{c}{2s} \\ \cline{3-6} 
                        &                                & PSNR            & MS-SSIM           & PSNR      & MS-SSIM    \\ \hline
Dither with Img. Grad.  & 51675                          & 27.15           & 0.9347            & 37.42     & 0.9934     \\ \hline
Color Field             & 77451                          & 30.21           & 0.9740            & 38.49     & 0.9968     \\ \hline
Init. with Fitted Ellipse     & 76881                          & 17.87           & 0.6942            & 36.94     & 0.9968     \\ \hline
Ker. Size = 4 (w/o train)     & 43627                          & 24.31           & 0.8911            & 35.12     & 0.9936     \\ \hline
Ker. Size = 4 (train)        & 44639                          & 29.53           & 0.9724            & 35.99     & 0.9945     \\ \hline
Final                   & 76881                          & 30.26           & 0.9752            & 38.73     & 0.9968     \\ \hline
\end{tabular}
}
\caption{\textbf{Ablation Study Results.} Quantitative evaluation of different strategies in our method. We compare initialization and 2s performance in terms of PSNR and MS-SSIM across various settings, including dithering with image gradients, using a color field, initializing with a fitted ellipse, and using different kernel sizes. 
%The final configuration achieves the best overall balance between initialization and fast convergence.
}
\label{tab:ablation}
\end{table}

\begin{figure}[ht!]
\begin{center}
\includegraphics[width=0.95\linewidth]{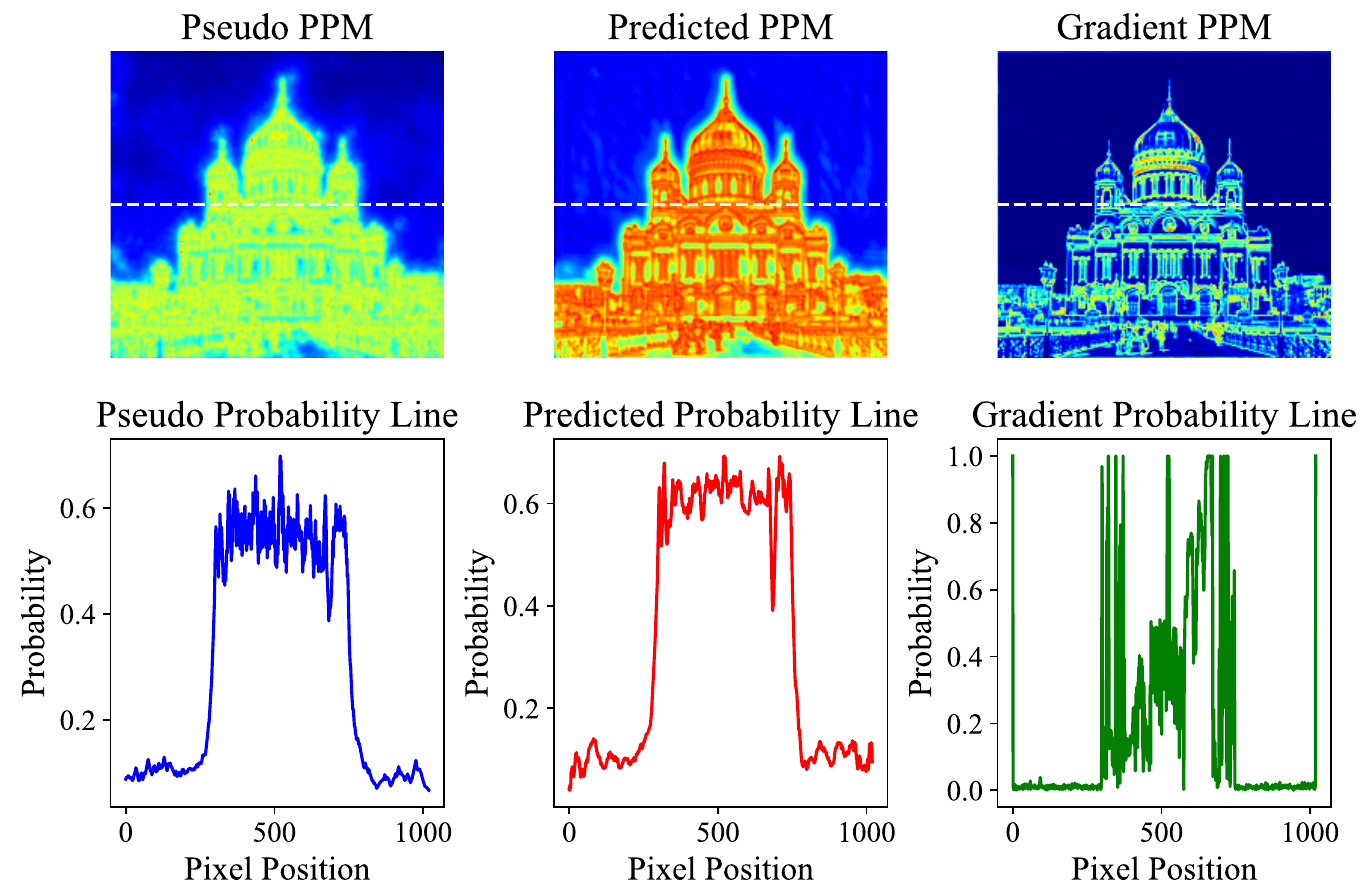}
\end{center}
\vspace{-0.3cm}
   \caption{\textbf{PPM Comparison.} Visualization of different Position Probability Maps (PPMs). The top row shows the pseudo PPM, predicted PPM, and gradient-based PPM. The bottom row presents their corresponding probability distributions along the indicated horizontal line. Our predicted PPM closely aligns with the pseudo PPM, while the gradient-based PPM exhibits abrupt transitions, highlighting the advantages of our learning-based approach.
}
\label{fig:probability_map}
\end{figure}

\noindent\textbf{Dithering using Image Gradient.}  
In \cref{sec:self_adaptive_gaussians_sample} we point out that gradient-based sampling is sub-optimal. The key issue, as observed in our experiments, is that high-entropy regions tend to have dense point distributions, while low-entropy regions are sparsely populated. However, at the boundaries between these regions, a smooth transition is necessary. Gradient-based probability maps rely solely on local information and fail to capture the overall density distribution. As shown in Fig.~\ref{fig:probability_map}, analyzing probability values along the horizontal center of the image reveals that both our method and the pseudo-probability distribution exhibit smooth transitions. In contrast, gradient-based maps introduce abrupt changes at high-low entropy boundaries, resulting in uneven density distributions and poor transitions. Essentially, a gradient-based probability map acts as an edge detector, leading to excessive sampling along edges while significantly reducing point density in smooth regions. This imbalance yields fewer total samples and, as \cref{tab:ablation} confirms, degrades rendering quality.

% In \cref{sec:self_adaptive_gaussians_sample} we point out that gradient-based sampling is sub-optimal. Experiments show that high-entropy regions naturally attract many points, while low-entropy regions attract few, so the probability map must change smoothly between them. Gradient maps, which rely only on local cues, ignore this global density and instead behave like edge detectors—over-sampling boundaries and under-sampling smooth areas. As Fig.~\ref{fig:probability_map} illustrates, our method and the pseudo-probability map trace a smooth profile across the image centre, whereas the gradient map jumps abruptly at high/low-entropy borders. This imbalance yields fewer total samples and, as \cref{tab:ablation} confirms, degrades rendering quality.

\noindent\textbf{Color Field.} 
In \cref{sec:gaussian_fields}, we proposed learning opacity and sampling colors from the image rather than directly predicting color as GI. This reduces the number of learned parameters from three (RGB with opacity) to one (opacity). On one hand, this approach does not compromise the model's expressive capability, as shown in Table~\ref{tab:ablation}. On the other hand, directly learning color slows down training and increases the likelihood of getting stuck in local minima. Please refer to the supplementary material for more details.

\noindent\textbf{Initialization with Fitted Ellipse.} 
As mentioned in \cref{sec:ellipse_fitting}, the fitted ellipse acts a reference for each primitive. So, why not use it directly for Gaussian initialization?  As shown in \cref{tab:ablation}, initializing Gaussians directly with the fitted ellipses can't yield reasonable scaling, rotation, and color. Additionally, at 2s, the PSNR of this method is lower than that obtained through network-based initialization.

\noindent\textbf{Kernel Size = 4.}  
A key feature of our algorithm is its ability to automatically determine the optimal number of Gaussians for a given image. But how can we achieve different compression ratios? As discussed in \cref{sec:dithering}, our method controls the number of generated points by setting a kernel size \( k \). Increasing \( k \) will reduce the number of generated points. Our model was trained with \( k = 3 \), achieving a well-optimized image representation. However, we can directly apply \( k = 4 \) to the same model without retraining. The corresponding rendering results are shown in \cref{tab:ablation} as ``Kernel size = 4 (w/o train)''. Similarly, we can fine-tune the model trained with \( k = 3 \) to adapt to \( k = 4 \), with the corresponding rendering performance listed as ``Kernel size = 4 (train)'' in \cref{tab:ablation}.  From the results, we observe that our method still produces a good result even without retraining, and fine-tuning significantly improves initialization quality. Meanwhile, the number of generated points decreases, effectively achieving a higher compression ratio.

%% file: sec/5_conclusion.tex
\section{Conclusion}
\label{sec:conclusion}

In this paper, we propose Instant-GI, a generalizable and self-adaptive image representation framework based on 2D Gaussian Splatting. Our method generates high-quality Gaussian representations through a single feedforward pass, followed by minimal fine-tuning to match or surpass the rendering performance of GI, which requires significantly longer training. Furthermore, our approach adaptively determines the number of Gaussians based on the image information entropy, ensuring consistent performance across different datasets.  Experimental results in the DIV2K and Kodak datasets show that, given the same number of Gaussians, Instant-GI achieves a rendering performance comparable to or surpass GI while significantly reducing training times. These findings underscore efficiency and practicality of our approach for high-quality image representation. For future work, we aim to further enhance rendering quality by exploring alternative initialization and optimization strategies. Our current pipeline incurs notable CPU overhead, particularly in Delaunay Triangulation, which could be optimized for better efficiency. Lastly, scaling our method to larger datasets may further improve its performance, contributing to a more robust image representation framework.

%% file: sec/6_acknowledge.tex
\section{Acknowledgements}
This work is supported by the Natural Science Foundation of China under Grant 62302174. The computation is completed in the HPC Platform of Huazhong University of Science and Technology. We also thank Farsee2 Technology Ltd for providing devices to support the validation of our method.

%% file: main.bbl
\begin{thebibliography}{50}
\providecommand{\natexlab}[1]{#1}
\providecommand{\url}[1]{\texttt{#1}}
\expandafter\ifx\csname urlstyle\endcsname\relax
  \providecommand{\doi}[1]{doi: #1}\else
  \providecommand{\doi}{doi: \begingroup \urlstyle{rm}\Url}\fi

\bibitem[Agustsson and Timofte(2017)]{Div2k}
Eirikur Agustsson and Radu Timofte.
\newblock Ntire 2017 challenge on single image super-resolution: Dataset and study.
\newblock In \emph{Proceedings of the IEEE conference on computer vision and pattern recognition workshops}, pages 126--135, 2017.

\bibitem[Barron et~al.(2021)Barron, Mildenhall, Tancik, Hedman, Martin-Brualla, and Srinivasan]{barron2021mip}
Jonathan~T Barron, Ben Mildenhall, Matthew Tancik, Peter Hedman, Ricardo Martin-Brualla, and Pratul~P Srinivasan.
\newblock Mip-nerf: A multiscale representation for anti-aliasing neural radiance fields.
\newblock In \emph{Proceedings of the IEEE/CVF International Conference on Computer Vision}, pages 5855--5864, 2021.

\bibitem[Barron et~al.(2023)Barron, Mildenhall, Verbin, Srinivasan, and Hedman]{barron2023zipnerf}
Jonathan~T. Barron, Ben Mildenhall, Dor Verbin, Pratul~P. Srinivasan, and Peter Hedman.
\newblock Zip-nerf: Anti-aliased grid-based neural radiance fields.
\newblock \emph{ICCV}, 2023.

\bibitem[Belhe et~al.(2023)Belhe, Gharbi, Fisher, Georgiev, Ramamoorthi, and Li]{belhe2023discontinuity}
Yash Belhe, Micha{\"e}l Gharbi, Matthew Fisher, Iliyan Georgiev, Ravi Ramamoorthi, and Tzu-Mao Li.
\newblock Discontinuity-aware 2d neural fields.
\newblock \emph{ACM Transactions on Graphics (TOG)}, 42\penalty0 (6):\penalty0 1--11, 2023.

\bibitem[Boris(1934)]{DelaunayTri}
N Boris.
\newblock Delaunay. sur la sphere vide.
\newblock \emph{Izvestia Akademia Nauk SSSR, VII Seria, Otdelenie Matematicheskii i Estestvennyka Nauk}, 7:\penalty0 793--800, 1934.

\bibitem[Chen et~al.(2024{\natexlab{a}})Chen, Li, Ye, Wang, Xie, Zhai, Wang, Liu, Bao, and Zhang]{chen2024pgsr}
Danpeng Chen, Hai Li, Weicai Ye, Yifan Wang, Weijian Xie, Shangjin Zhai, Nan Wang, Haomin Liu, Hujun Bao, and Guofeng Zhang.
\newblock Pgsr: Planar-based gaussian splatting for efficient and high-fidelity surface reconstruction.
\newblock \emph{IEEE Transactions on Visualization and Computer Graphics}, 2024{\natexlab{a}}.

\bibitem[Chen et~al.(2021{\natexlab{a}})Chen, He, Wang, Ren, Lim, and Shrivastava]{chen2021nerv}
Hao Chen, Bo He, Hanyu Wang, Yixuan Ren, Ser~Nam Lim, and Abhinav Shrivastava.
\newblock Nerv: Neural representations for videos.
\newblock \emph{Advances in Neural Information Processing Systems}, 34:\penalty0 21557--21568, 2021{\natexlab{a}}.

\bibitem[Chen et~al.(2023{\natexlab{a}})Chen, Gwilliam, Lim, and Shrivastava]{chen2023hnerv}
Hao Chen, Matthew Gwilliam, Ser-Nam Lim, and Abhinav Shrivastava.
\newblock Hnerv: A hybrid neural representation for videos.
\newblock In \emph{Proceedings of the IEEE/CVF Conference on Computer Vision and Pattern Recognition}, pages 10270--10279, 2023{\natexlab{a}}.

\bibitem[Chen et~al.(2021{\natexlab{b}})Chen, Liu, and Wang]{LIIF}
Yinbo Chen, Sifei Liu, and Xiaolong Wang.
\newblock Learning continuous image representation with local implicit image function.
\newblock In \emph{Proceedings of the IEEE/CVF conference on computer vision and pattern recognition}, pages 8628--8638, 2021{\natexlab{b}}.

\bibitem[Chen et~al.(2024{\natexlab{b}})Chen, Chen, Zhang, Wang, Yang, Wang, Cai, Yang, Liu, and Lin]{chen2024gaussianeditor}
Yiwen Chen, Zilong Chen, Chi Zhang, Feng Wang, Xiaofeng Yang, Yikai Wang, Zhongang Cai, Lei Yang, Huaping Liu, and Guosheng Lin.
\newblock Gaussianeditor: Swift and controllable 3d editing with gaussian splatting.
\newblock In \emph{Proceedings of the IEEE/CVF conference on computer vision and pattern recognition}, pages 21476--21485, 2024{\natexlab{b}}.

\bibitem[Chen et~al.(2023{\natexlab{b}})Chen, Li, Song, Chen, Yu, Yuan, and Xu]{Neurbf}
Zhang Chen, Zhong Li, Liangchen Song, Lele Chen, Jingyi Yu, Junsong Yuan, and Yi Xu.
\newblock Neurbf: A neural fields representation with adaptive radial basis functions.
\newblock In \emph{Proceedings of the IEEE/CVF International Conference on Computer Vision}, pages 4182--4194, 2023{\natexlab{b}}.

\bibitem[Dong et~al.(2015)Dong, Loy, He, and Tang]{dong2015image}
Chao Dong, Chen~Change Loy, Kaiming He, and Xiaoou Tang.
\newblock Image super-resolution using deep convolutional networks.
\newblock \emph{IEEE transactions on pattern analysis and machine intelligence}, 38\penalty0 (2):\penalty0 295--307, 2015.

\bibitem[Fitzgibbon et~al.(1996)Fitzgibbon, Fisher, et~al.]{fitEllipse}
Andrew~W Fitzgibbon, Robert~B Fisher, et~al.
\newblock \emph{A buyer's guide to conic fitting}.
\newblock Citeseer, 1996.

\bibitem[Franchini et~al.(2019)Franchini, Cavicchioli, and Hu]{DitheringGPU}
Giorgia Franchini, Roberto Cavicchioli, and Jia~Cheng Hu.
\newblock Stochastic floyd-steinberg dithering on gpu: image quality and processing time improved.
\newblock In \emph{2019 Fifth International Conference on Image Information Processing (ICIIP)}, pages 1--6, 2019.

\bibitem[Hu et~al.(2024)Hu, Xia, Chen, Yang, and Zhang]{GaussianSR}
Jintong Hu, Bin Xia, Bin Chen, Wenming Yang, and Lei Zhang.
\newblock Gaussiansr: High fidelity 2d gaussian splatting for arbitrary-scale image super-resolution, 2024.

\bibitem[Hu et~al.(2019)Hu, Mu, Zhang, Wang, Tan, and Sun]{hu2019meta}
Xuecai Hu, Haoyuan Mu, Xiangyu Zhang, Zilei Wang, Tieniu Tan, and Jian Sun.
\newblock Meta-sr: A magnification-arbitrary network for super-resolution.
\newblock In \emph{Proceedings of the IEEE/CVF conference on computer vision and pattern recognition}, pages 1575--1584, 2019.

\bibitem[Huang et~al.(2024)Huang, Yu, Chen, Geiger, and Gao]{huang20242d}
Binbin Huang, Zehao Yu, Anpei Chen, Andreas Geiger, and Shenghua Gao.
\newblock 2d gaussian splatting for geometrically accurate radiance fields.
\newblock In \emph{ACM SIGGRAPH 2024 conference papers}, pages 1--11, 2024.

\bibitem[Jaganathan et~al.(2024)Jaganathan, Huang, Irshad, Jampani, Raj, and Kira]{jaganathan2024ice}
Vishnu Jaganathan, Hannah~Hanyun Huang, Muhammad~Zubair Irshad, Varun Jampani, Amit Raj, and Zsolt Kira.
\newblock Ice-g: Image conditional editing of 3d gaussian splats.
\newblock \emph{arXiv preprint arXiv:2406.08488}, 2024.

\bibitem[Kerbl et~al.(2023)Kerbl, Kopanas, Leimk{\"u}hler, and Drettakis]{3DGS}
Bernhard Kerbl, Georgios Kopanas, Thomas Leimk{\"u}hler, and George Drettakis.
\newblock 3d gaussian splatting for real-time radiance field rendering.
\newblock \emph{ACM Trans. Graph.}, 42\penalty0 (4):\penalty0 139--1, 2023.

\bibitem[Kingma and Ba(2014)]{Adam}
Diederik~P Kingma and Jimmy Ba.
\newblock Adam: A method for stochastic optimization.
\newblock \emph{arXiv preprint arXiv:1412.6980}, 2014.

\bibitem[Klocek et~al.(2019)Klocek, Maziarka, Wo{\l}czyk, Tabor, Nowak, and {\'S}mieja]{klocek2019hypernetwork}
Sylwester Klocek, {\L}ukasz Maziarka, Maciej Wo{\l}czyk, Jacek Tabor, Jakub Nowak, and Marek {\'S}mieja.
\newblock Hypernetwork functional image representation.
\newblock In \emph{International Conference on Artificial Neural Networks}, pages 496--510. Springer, 2019.

\bibitem[Kodak(1999)]{Kodak}
Kodak.
\newblock Kodak lossless true color image suite, 1999.

\bibitem[Kong et~al.(2023)Kong, Dong, Ge, Li, and Pan]{Kong_2023_CVPR}
Lingshun Kong, Jiangxin Dong, Jianjun Ge, Mingqiang Li, and Jinshan Pan.
\newblock Efficient frequency domain-based transformers for high-quality image deblurring.
\newblock In \emph{Proceedings of the IEEE/CVF Conference on Computer Vision and Pattern Recognition (CVPR)}, pages 5886--5895, 2023.

\bibitem[Kuznetsov(2021)]{kuznetsov2021neumip}
Alexandr Kuznetsov.
\newblock Neumip: Multi-resolution neural materials.
\newblock \emph{ACM Transactions on Graphics (ToG)}, 40\penalty0 (4), 2021.

\bibitem[Li et~al.(2022)Li, Dai, Li, Zou, and Xia]{li2022adaptive}
Hongwei Li, Tao Dai, Yiming Li, Xueyi Zou, and Shu-Tao Xia.
\newblock Adaptive local implicit image function for arbitrary-scale super-resolution.
\newblock In \emph{2022 IEEE International Conference on Image Processing (ICIP)}, pages 4033--4037. IEEE, 2022.

\bibitem[Li et~al.(2023)Li, Wang, and Tseng]{li2023gaussiandiffusion}
Xinhai Li, Huaibin Wang, and Kuo-Kun Tseng.
\newblock Gaussiandiffusion: 3d gaussian splatting for denoising diffusion probabilistic models with structured noise.
\newblock \emph{arXiv preprint arXiv:2311.11221}, 2023.

\bibitem[Lim et~al.(2017)Lim, Son, Kim, Nah, and Mu~Lee]{lim2017enhanced}
Bee Lim, Sanghyun Son, Heewon Kim, Seungjun Nah, and Kyoung Mu~Lee.
\newblock Enhanced deep residual networks for single image super-resolution.
\newblock In \emph{Proceedings of the IEEE conference on computer vision and pattern recognition workshops}, pages 136--144, 2017.

\bibitem[Liu et~al.(2024)Liu, Wang, Hu, Shen, Ye, Zang, Cao, Li, and Liu]{MVSGaussian}
Tianqi Liu, Guangcong Wang, Shoukang Hu, Liao Shen, Xinyi Ye, Yuhang Zang, Zhiguo Cao, Wei Li, and Ziwei Liu.
\newblock Mvsgaussian: Fast generalizable gaussian splatting reconstruction from multi-view stereo.
\newblock In \emph{European Conference on Computer Vision}, pages 37--53. Springer, 2024.

\bibitem[Liu et~al.(2022)Liu, Mao, Wu, Feichtenhofer, Darrell, and Xie]{liu2022convnet}
Zhuang Liu, Hanzi Mao, Chao-Yuan Wu, Christoph Feichtenhofer, Trevor Darrell, and Saining Xie.
\newblock A convnet for the 2020s.
\newblock In \emph{Proceedings of the IEEE/CVF conference on computer vision and pattern recognition}, pages 11976--11986, 2022.

\bibitem[Loshchilov and Hutter(2016)]{loshchilov2016sgdr}
Ilya Loshchilov and Frank Hutter.
\newblock Sgdr: Stochastic gradient descent with warm restarts.
\newblock \emph{arXiv preprint arXiv:1608.03983}, 2016.

\bibitem[Lu et~al.(2024)Lu, Yu, Xu, Xiangli, Wang, Lin, and Dai]{Scaffold-GS}
Tao Lu, Mulin Yu, Linning Xu, Yuanbo Xiangli, Limin Wang, Dahua Lin, and Bo Dai.
\newblock Scaffold-gs: Structured 3d gaussians for view-adaptive rendering.
\newblock In \emph{Proceedings of the IEEE/CVF Conference on Computer Vision and Pattern Recognition}, pages 20654--20664, 2024.

\bibitem[Martel et~al.(2021)Martel, Lindell, Lin, Chan, Monteiro, and Wetzstein]{martel2021acorn}
Julien~NP Martel, David~B Lindell, Connor~Z Lin, Eric~R Chan, Marco Monteiro, and Gordon Wetzstein.
\newblock Acorn: Adaptive coordinate networks for neural scene representation.
\newblock \emph{arXiv preprint arXiv:2105.02788}, 2021.

\bibitem[Mildenhall et~al.(2020)Mildenhall, Srinivasan, Tancik, Barron, Ramamoorthi, and Ng]{NeRF}
B Mildenhall, PP Srinivasan, M Tancik, JT Barron, R Ramamoorthi, and R Ng.
\newblock Nerf: Representing scenes as neural radiance fields for view synthesis.
\newblock In \emph{European conference on computer vision}, 2020.

\bibitem[Mou et~al.(2022)Mou, Wang, and Zhang]{mou2022deep}
Chong Mou, Qian Wang, and Jian Zhang.
\newblock Deep generalized unfolding networks for image restoration.
\newblock In \emph{Proceedings of the IEEE/CVF conference on computer vision and pattern recognition}, pages 17399--17410, 2022.

\bibitem[M{\"u}ller et~al.(2022)M{\"u}ller, Evans, Schied, and Keller]{I-NGP}
Thomas M{\"u}ller, Alex Evans, Christoph Schied, and Alexander Keller.
\newblock Instant neural graphics primitives with a multiresolution hash encoding.
\newblock \emph{ACM transactions on graphics (TOG)}, 41\penalty0 (4):\penalty0 1--15, 2022.

\bibitem[Saragadam et~al.(2023)Saragadam, LeJeune, Tan, Balakrishnan, Veeraraghavan, and Baraniuk]{Wire}
Vishwanath Saragadam, Daniel LeJeune, Jasper Tan, Guha Balakrishnan, Ashok Veeraraghavan, and Richard~G Baraniuk.
\newblock Wire: Wavelet implicit neural representations.
\newblock In \emph{Proceedings of the IEEE/CVF Conference on Computer Vision and Pattern Recognition}, pages 18507--18516, 2023.

\bibitem[Sitzmann et~al.(2020)Sitzmann, Martel, Bergman, Lindell, and Wetzstein]{SIREN}
Vincent Sitzmann, Julien Martel, Alexander Bergman, David Lindell, and Gordon Wetzstein.
\newblock Implicit neural representations with periodic activation functions.
\newblock \emph{Advances in neural information processing systems}, 33:\penalty0 7462--7473, 2020.

\bibitem[Takikawa et~al.(2023)Takikawa, M{\"u}ller, Nimier-David, Evans, Fidler, Jacobson, and Keller]{takikawa2023compact}
Towaki Takikawa, Thomas M{\"u}ller, Merlin Nimier-David, Alex Evans, Sanja Fidler, Alec Jacobson, and Alexander Keller.
\newblock Compact neural graphics primitives with learned hash probing.
\newblock In \emph{SIGGRAPH Asia 2023 Conference Papers}, pages 1--10, 2023.

\bibitem[Tancik et~al.(2020)Tancik, Srinivasan, Mildenhall, Fridovich-Keil, Raghavan, Singhal, Ramamoorthi, Barron, and Ng]{tancik2020fourier}
Matthew Tancik, Pratul Srinivasan, Ben Mildenhall, Sara Fridovich-Keil, Nithin Raghavan, Utkarsh Singhal, Ravi Ramamoorthi, Jonathan Barron, and Ren Ng.
\newblock Fourier features let networks learn high frequency functions in low dimensional domains.
\newblock \emph{Advances in neural information processing systems}, 33:\penalty0 7537--7547, 2020.

\bibitem[Tang et~al.(2023)Tang, Ren, Zhou, Liu, and Zeng]{tang2023dreamgaussian}
Jiaxiang Tang, Jiawei Ren, Hang Zhou, Ziwei Liu, and Gang Zeng.
\newblock Dreamgaussian: Generative gaussian splatting for efficient 3d content creation.
\newblock \emph{arXiv preprint arXiv:2309.16653}, 2023.

\bibitem[Vaidyanathan et~al.(2023)Vaidyanathan, Salvi, Wronski, Akenine-M{\"o}ller, Ebelin, and Lefohn]{vaidyanathan2023random}
Karthik Vaidyanathan, Marco Salvi, Bartlomiej Wronski, Tomas Akenine-M{\"o}ller, Pontus Ebelin, and Aaron Lefohn.
\newblock Random-access neural compression of material textures.
\newblock \emph{arXiv preprint arXiv:2305.17105}, 2023.

\bibitem[Waczy{\'n}ska et~al.(2024)Waczy{\'n}ska, Szczepanik, Borycki, Tadeja, Bohn{\'e}, and Spurek]{Mirage}
Joanna Waczy{\'n}ska, Tomasz Szczepanik, Piotr Borycki, S{\l}awomir Tadeja, Thomas Bohn{\'e}, and Przemys{\l}aw Spurek.
\newblock Mirage: Editable 2d images using gaussian splatting.
\newblock \emph{arXiv preprint arXiv:2410.01521}, 2024.

\bibitem[Wang et~al.(2021)Wang, Liu, Liu, Theobalt, Komura, and Wang]{wang2021neus}
Peng Wang, Lingjie Liu, Yuan Liu, Christian Theobalt, Taku Komura, and Wenping Wang.
\newblock Neus: Learning neural implicit surfaces by volume rendering for multi-view reconstruction.
\newblock \emph{arXiv preprint arXiv:2106.10689}, 2021.

\bibitem[Wang et~al.(2024)Wang, Yi, Wu, Zhao, Chen, and Zhang]{wang2024view}
Yuxuan Wang, Xuanyu Yi, Zike Wu, Na Zhao, Long Chen, and Hanwang Zhang.
\newblock View-consistent 3d editing with gaussian splatting.
\newblock In \emph{European Conference on Computer Vision}, pages 404--420. Springer, 2024.

\bibitem[Wu et~al.(2023)Wu, Ni, and Zhang]{wu2023learning}
Hanlin Wu, Ning Ni, and Libao Zhang.
\newblock Learning dynamic scale awareness and global implicit functions for continuous-scale super-resolution of remote sensing images.
\newblock \emph{IEEE Transactions on Geoscience and Remote Sensing}, 61:\penalty0 1--15, 2023.

\bibitem[Yu et~al.(2024)Yu, Sattler, and Geiger]{yu2024gaussian}
Zehao Yu, Torsten Sattler, and Andreas Geiger.
\newblock Gaussian opacity fields: Efficient adaptive surface reconstruction in unbounded scenes.
\newblock \emph{ACM Transactions on Graphics (TOG)}, 43\penalty0 (6):\penalty0 1--13, 2024.

\bibitem[Zhang et~al.(2024{\natexlab{a}})Zhang, Fang, Shrestha, Liang, Long, and Tan]{zhang2024rade}
Baowen Zhang, Chuan Fang, Rakesh Shrestha, Yixun Liang, Xiaoxiao Long, and Ping Tan.
\newblock Rade-gs: Rasterizing depth in gaussian splatting.
\newblock \emph{arXiv preprint arXiv:2406.01467}, 2024{\natexlab{a}}.

\bibitem[Zhang et~al.(2024{\natexlab{b}})Zhang, Ge, Xu, He, Wang, Qin, Lu, Geng, and Zhang]{GaussianImage}
Xinjie Zhang, Xingtong Ge, Tongda Xu, Dailan He, Yan Wang, Hongwei Qin, Guo Lu, Jing Geng, and Jun Zhang.
\newblock Gaussianimage: 1000 fps image representation and compression by 2d gaussian splatting.
\newblock In \emph{European Conference on Computer Vision}, pages 327--345. Springer, 2024{\natexlab{b}}.

\bibitem[Zhang et~al.(2024{\natexlab{c}})Zhang, Kuznetsov, Jindal, Chen, Sochenov, Kaplanyan, and Sun]{ImageGS}
Yunxiang Zhang, Alexandr Kuznetsov, Akshay Jindal, Kenneth Chen, Anton Sochenov, Anton Kaplanyan, and Qi Sun.
\newblock Image-gs: Content-adaptive image representation via 2d gaussians.
\newblock \emph{arXiv preprint arXiv:2407.01866}, 2024{\natexlab{c}}.

\bibitem[Zhou et~al.(2024)Zhou, Zhang, and Liu]{DiffGS}
Junsheng Zhou, Weiqi Zhang, and Yu-Shen Liu.
\newblock Diffgs: Functional gaussian splatting diffusion.
\newblock In \emph{Advances in Neural Information Processing Systems (NeurIPS)}, 2024.

\end{thebibliography}


\begin{thebibliography}{5}
\providecommand{\natexlab}[1]{#1}
\providecommand{\url}[1]{\texttt{#1}}
\expandafter\ifx\csname urlstyle\endcsname\relax
  \providecommand{\doi}[1]{doi: #1}\else
  \providecommand{\doi}{doi: \begingroup \urlstyle{rm}\Url}\fi

\bibitem[Fitzgibbon et~al.(1996)Fitzgibbon, Fisher, et~al.]{fitEllipse}
Andrew~W Fitzgibbon, Robert~B Fisher, et~al.
\newblock \emph{A buyer's guide to conic fitting}.
\newblock Citeseer, 1996.

\bibitem[Franchini et~al.(2019)Franchini, Cavicchioli, and Hu]{DitheringGPU}
Giorgia Franchini, Roberto Cavicchioli, and Jia~Cheng Hu.
\newblock Stochastic floyd-steinberg dithering on gpu: image quality and processing time improved.
\newblock In \emph{2019 Fifth International Conference on Image Information Processing (ICIIP)}, pages 1--6, 2019.

\bibitem[Kerbl et~al.(2023)Kerbl, Kopanas, Leimk{\"u}hler, and Drettakis]{3DGS}
Bernhard Kerbl, Georgios Kopanas, Thomas Leimk{\"u}hler, and George Drettakis.
\newblock 3d gaussian splatting for real-time radiance field rendering.
\newblock \emph{ACM Trans. Graph.}, 42\penalty0 (4):\penalty0 139--1, 2023.

\bibitem[Lu et~al.(2024)Lu, Yu, Xu, Xiangli, Wang, Lin, and Dai]{Scaffold-GS}
Tao Lu, Mulin Yu, Linning Xu, Yuanbo Xiangli, Limin Wang, Dahua Lin, and Bo Dai.
\newblock Scaffold-gs: Structured 3d gaussians for view-adaptive rendering.
\newblock In \emph{Proceedings of the IEEE/CVF Conference on Computer Vision and Pattern Recognition}, pages 20654--20664, 2024.

\bibitem[Zhang et~al.(2024)Zhang, Kuznetsov, Jindal, Chen, Sochenov, Kaplanyan, and Sun]{ImageGS}
Yunxiang Zhang, Alexandr Kuznetsov, Akshay Jindal, Kenneth Chen, Anton Sochenov, Anton Kaplanyan, and Qi Sun.
\newblock Image-gs: Content-adaptive image representation via 2d gaussians.
\newblock \emph{arXiv preprint arXiv:2407.01866}, 2024.

\end{thebibliography}
